\documentclass[conference]{IEEEtran}
\IEEEoverridecommandlockouts
\usepackage{cite}
\usepackage{amsmath,amssymb,amsfonts}
\usepackage{graphicx}
\usepackage{textcomp}
\usepackage{xcolor}
\usepackage{times}
\usepackage{soul}
\usepackage{url}
\usepackage[utf8]{inputenc}
\usepackage[small]{caption}
\usepackage{graphicx}
\usepackage{booktabs}
\usepackage{algorithm}
\usepackage{amsthm}
\newtheorem{definition}{Definition}
\usepackage[bookmarks=false]{hyperref}
\usepackage{mathtools}
\usepackage{algpseudocode}
\usepackage{multirow}
\usepackage{subcaption}
\usepackage{color}
\usepackage{subcaption}
\usepackage{placeins}

\newtheorem{theorem}{Theorem}

\DeclarePairedDelimiter\ceil{\lceil}{\rceil}
\DeclarePairedDelimiter\floor{\lfloor}{\rfloor}
\def\BibTeX{{\rm B\kern-.05em{\sc i\kern-.025em b}\kern-.08em
    T\kern-.1667em\lower.7ex\hbox{E}\kern-.125emX}}
\begin{document}

\title{Rank-Based Multi-task Learning For Fair Regression\\
}


\author{\IEEEauthorblockN{Chen Zhao}
\IEEEauthorblockA{\textit{Department of Computer Science} \\
\textit{The University of Texas at Dallas}\\
Dallas Texas, USA \\
chen.zhao@utdallas.edu}
\and
\IEEEauthorblockN{Feng Chen}
\IEEEauthorblockA{\textit{Department of Computer Science} \\
\textit{The University of Texas at Dallas}\\
Dallas Texas, USA \\
feng.chen@utdallas.edu}
}

\maketitle
\vspace{-20mm}
\begin{abstract}

In this work, we develop a novel fairness learning approach for multi-task regression models based on a biased training dataset, using a popular rank-based non-parametric independence test, i.e., Mann Whitney U statistic, for measuring the dependency between target variable and protected variables. To solve this learning problem efficiently, we first reformulate the problem as a new non-convex optimization problem, in which a non-convex constraint is defined based on group-wise ranking functions of individual objects. We then develop an efficient model-training algorithm based on the framework of non-convex alternating direction method of multipliers (NC-ADMM), in which one of the main challenges is to implement an efficient projection oracle to the preceding non-convex set defined based on ranking functions. Through the extensive  experiments on both synthetic and real-world datasets, we validated the out-performance of our new approach against several state-of-the-art competitive methods on several popular metrics relevant to fairness learning. 
\end{abstract}

\begin{IEEEkeywords}
sum rank, Mann Whitney U statistic, multi-task learning, NC-ADMM, fairness
\end{IEEEkeywords}

\section{Introduction}
Data-driven and big data technologies, nowadays, have advanced many complex domains such as healthcare, finance, social science, etc. With the development and increment of data, it is necessary to extract the potential and significant knowledge and unveil the messages hidden behind using data analysis. In data mining and machine learning, biased historical data are often learned and used to train statistic predictive model. Depending on application field, even though predictive models and computing process is fair, biased training data or data containing discrimination may lead to results with undesirability, inaccuracy and even illegality. For example, in recent years, there have been a number of news articles that discuss the concerns of bias and discrimination on crime forecasting. 911 call data was used to predict the locations of crimes by the Seattle Police Department in 2016, but was late dropped due to potential racial bias in this data \cite{SeattlePolice}.

Non-discrimination can be defined as follows: (1) people that are similar in terms of non-sensitive characteristics should receive similar predictions, and (2) differences in predictions across groups of people can only be as large as justified by non-sensitive characteristics\cite{Zliobaite2015}. The first condition is related to direct discrimination. For example, a hotel turns a customer away due to disability. The second condition ensures that there is no indirect discrimination, also referred to as redlining. For example, one is treated in the same way as everybody else, but it has a different and worse effect because one's gender, race or other sensitive characters. The Equality Act \footnote{The U.S. Civil Rights Act, July 2, 1964} calls these characters as protected characteristics. In this paper, we consider fair machine learning algorithms that removes the discrimination impact arising from the strong dependency of the protected variable upon predicted outputs.

To the best of our knowledge, unfortunately, most of existing fairness-aware machine learning algorithms, in terms of underlying generic learning methods, focus on solving single-task learning (STL) and independent-task learning (ITL) problems. However, when problem requires more than one group (i.e. task), both STL and ITL approaches have limitations. While the STL learns a function that is common across entire data, it may learn a model representing the largest group. The ITL independently learns a different function for each group, but it may overfit minority groups \cite{Oneto2018}. A common methodology to overcome such limitations is multi-task learning (MTL). MTL jointly learns a shared model for all groups as well as a specific model per group and leverages information between them. Besides, in contrast to single-task problem, we added the $\ell_{2,1}$-norm regularization \cite{Liu2009} that encourages multiple predictors from various tasks to share similar parameter sparsity patterns. It penalizes the sum of the $\ell_2$-norms of the blocks of coefficients associated with each attribute across tasks, leading to an $\ell_{2,1}$-norm regularized non-smooth convex loss function.

In this paper, we study fairness in multi-task regression models, which have been shown sensitive to discrimination that exists in historical data \cite{Oneto2018}. Although a number of methods have been proposed for classification models, limited methods have been designed for regression models. The overall idea of these methods is to train regression models such that their predictions are un-correlated with the predefined protected attributes (e.g., race and gender), based on measures such as mean difference \cite{Calders2013, Kamiran2010} and mutual information \cite{Fukuchi2013}. However, these methods have two main limitations: \textit{1)~Limited capability in eliminating independence rather than correlations}. These methods are incapable of eliminating the independence rather than correlations between predictions and protected attributes. \textit{2)~Lack of support for multi-task regression models.} These methods fail to provide bias-free predictions for multi-task regression models, i.e, controlling bias across multiple tasks simultaneously. 

\begin{figure}[h!]
    \centering
    \includegraphics[width=0.5\textwidth]{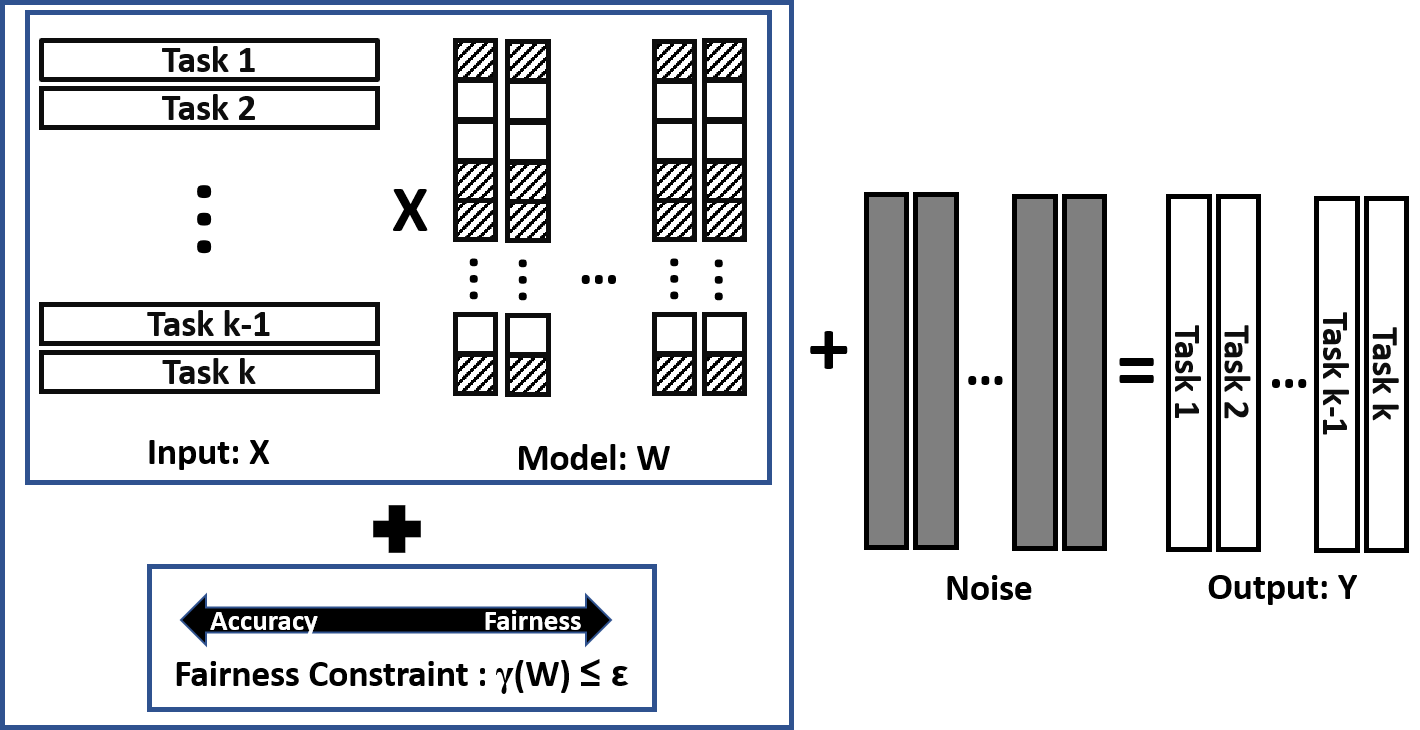}
    \caption{An overview of our proposed multi-task model with joint feature learning, where $Z$ denotes the protected/sensitive attribute, $\gamma(W)$ refers to a discrimination measure function, and $\epsilon$ is a predefined small threshold to account for a degree of randomness in the decision making process and sampling. The constraint here is considered to ensure that there is no indirect discrimination on the prediction model.}
    \label{fig:overview}
    \vspace{-7mm}
\end{figure}

Since history data may be collected from various sources or has dependency effect on socially protected attributes \cite{Calders2013}, as stated in Fig.\ref{fig:overview}, using biased training data to train the multi-task model for regression and expecting fairness output that requires us to restrict our model into fairness constraints in which the dependency effect of protected variables to prediction is eliminated. As is well known, there is a trade-off between fairness and accuracy, which implies that a perfect fair optimizer does not always yield a useful prediction \cite{Komiyama2018}.

We thus started by dividing data into two partitions according to the binary protected attribute and then sorted by target values in an ascending order. According to the order, each observation was assigned with a rank number which is a positive integer starting from 1. To control the dependency effect of the protected attribute on predictions, for example, African American on crime rate, a non-convex constraint based on limiting the sum rank of binary partitions was applied. This constraint representing independence was derived from restricting the \textit{U} value to a fairness level using non-parametric Mann Whitney \textit{U} statistic test. In summary, the main contributions of this paper are listed:
\begin{enumerate}
    \item We presented the first-known approach to the fairness learning problem in multi-task regression models based on a popular non-parametric independence statistic test, i.e., the Mann Whitney \textit{U} statistic. Our approach enables the eliminations of bias in predictions across multiple tasks simultaneously that are free of distribution assumptions.
    
    \item We reformulated the fairness learning problem as a new non-convex optimization problem, in which a non-convex constraint is defined based on group-wise ranking functions of individual objects. We developed efficient algorithms to solve the reformulated problem based on the framework of non-convex alternating direction method of multipliers (NC-ADMM). 
    
    \item We validate the performance of our proposed approach through the extensive experiments based on four synthetic and four real-world datasets. We conduct a performance comparison between our proposed approach and the existing counterparts in terms of several popular metrics relevant to fairness. 
\end{enumerate} 
Reproducibility: The implementation of our proposed method and datasets are available via the link: \url{https://bit.ly/2Xw7T2S}.

\section{Related Work}
In recent years, researches involving processing biased data became increasingly significant. Fairness-aware in data mining is classified into unfairness discovery and prevention. Based on the taxonomy by tasks, prevention on unfairness (bias) can be further categorized to classification, regression \cite{Calders2013,Fukuchi2013,Berk2017,Komiyama2018,PerezSuay2017}, clustering \cite{Gondek2004,Gondek2005}, recommendation \cite{Kamishima2016,Kamishima2017,Ashudeep2018} and dimension reduction \cite{Bolukbasi2016}. Majority of works in this field, however, is concentrated on data classification. According to approaches studied in fairness, bias-prevention in classification subcategorized into pre-processing\cite{Feldman2015}, in-processing \cite{Zafar2017, Fish2015, Zafar2015} and post-processing \cite{Hardt2016,Kamishima2018,Hajian2015}. The most recent work \cite{Agarwal2018} and \cite{Oneto2018} developed new approaches resulting in increasing the binary classification accuracy through reduction of fair classification to a sequence of cost-sensitive problem and application of multi-tasks technique with convex fairness constraints, respectively.

Even though techniques for unfairness prevention on classification were well developed, limited methods have been designed for regression models and the problem on regression is more challenging. Because (1) instead of evaluating the correlation between two categorical attributes, regression aims to assess the the correlation on the categorical protected attribute and continuous target variable; (2) in classification the goal of modification led to the change of one class label into another, however in a regression task, fairness learning allows the continuous character of targets for a continuous range of potential changes. \cite{Calders2013} first controlled bias in a regression model by restricting the mean difference in predictions on several data strata divided using the propensity scoring method from statistics. Furthermore, \cite{Fukuchi2013} proposed a framework involving $\eta$-neutrality in which to use a maximum likelihood estimation for learning probabilistic models. Besides, \cite{Berk2017} and \cite{Komiyama2018} recently came up with convex and non-convex optimization frameworks for fairness regression. 

\begin{table*}[h]
    \centering
    \captionof{table}{Comparison of Statistic Independent Tests for Single Binary Protected Variable $Z$}
    \label{tab:ind tests}
    \begin{tabular}{|c|c|c|c|}
        \hline
        Independent Test & Nature of $Z$ & Nature of $Y$ & Distribution Assumption \\
        \hline
        Student t-test & $Z$ with 2 levels (independent groups) & numeric & normal\\
        \hline
        Mann Whitney \textit{U} & $Z$ with 2 levels (independent groups) & numeric, ordinal & non-parametric\\
        \hline
        Chi-square test & $Z$ with 2 levels (independent groups) & categorical & non-parametric\\
        \hline
        Fisher's exact test & $Z$ with 2 levels (independent groups) & categorical & non-parametric\\
        \hline
        Paired t-test & $Z$ with 2 levels (dependent/matched groups) & numeric & normal\\
        \hline
        Wilcoxon signed-rank & $Z$ with 2 levels (dependent/matched groups) & numeric, ordinal & non-parametric\\
        \hline
        McNemar & $Z$ with 2 levels (dependent/matched groups) & categorical & non-parametric\\
        \hline
    \end{tabular}
\vspace{-5mm}
\end{table*}

To the best of our knowledge, the overall idea of these methods is to train regression models such that their predictions are un-correlated with the predefined protected attributes (e.g., race and gender), based on measures such as either mean difference, mutual information or correlation coefficient. Unlike existing approaches, our proposed method (1) provides prediction under for training data under no distribution assumption using multi-task learning for regression and (2) enables strict control of fairness over tasks by posing the fairness as an explicit optimization constraint.

\section{Rank Based Independence Test}
Intuitively, an attribute effects the target variable if one depends on the other. Strong dependency indicates strong effects. We are now looking for an independent test such that it measures the dependent effect of the binary protected attribute $Z$ (e.g. race) on numeric targets $Y$ (e.g. crime count). Although there are a number of statistic independent tests for numeric protected variables, in this paper, we only focus on non-paired/matched binary $Z$. Table \ref{tab:ind tests} lists several general guidelines for choosing a statistic independent test for single binary protected variable $Z$. Although Student’s t-test is widely applied to test how two samples significantly differ from each other. One major limitation for t-test is that samples being compared should follow a normal distribution. Contingency table associated with Chi-square and Fisher's exact are two alternative  tests. However, both of these two tests require that the nature of $Y$ is categorical rather numeric. 

To overcome such limitations, a popular non-parametric test to compare outcomes between two independent partitions (i.e. \textit{A} and \textit{B}) is the Mann Whitney \textit{U} test (a.k.a Wilcoxon Rank Sum Test or Mann Whitney Wilcoxon Test). This statistic test is used to test whether two samples are likely to derive from the same population. In other words, it compares differences between two independent partitions when the dependent variable is either ordinal or continuous, but not normally distributed. For example, one could use the Mann-Whitney \textit{U} test to understand whether attitudes towards crime discrimination, where attitudes are measured on an ordinal scale, differ based on the protected attribute. 

Suppose we have a data $\mathcal{D}$ which includes a sample of $|\mathcal{D}_A|$ observations $\{y^A_1,y^A_2,...,y^A_{|\mathcal{D}_A|}\}$ in partition \textit{A} (the superscript denotes the partition of the element), and a sample of $|\mathcal{D}_B|$ observations $\{y^B_1,y^B_2,...,y^B_{|\mathcal{D}_B|}\}$ in partition \textit{B}, where $\{A,B\}\in Z$, $y_i\in\mathbb{R}$ and the binary variable $Z$ decides two partitions. The Mann Whitney \textit{U} test is based on a comparison of every observation $y^A_i$ in partition \textit{A} with every observation $y^B_i$ in partition \textit{B}. The total number of pairwise comparisons that can be made is $|\mathcal{D}_A|\times|\mathcal{D}_B|$. The commonly stated hypotheses for Mann Whitney \textit{U} test are:
\begin{align*}
    &{\small H_0: \text{The distributions of $Y$ in the two partitions are the same.}}\\
    &{\small H_1: \text{The distributions of $Y$ in the two partitions are not the same.}}
\end{align*}
That is, under the null hypothesis, the probability of an observation from one partition $A$ exceeding an observation from the second partition $B$ equals the probability of an observation from $B$ exceeding an observation from $A$, i.e. $P(y|\mathcal{D}_A)=P(y|\mathcal{D}_B)=0.5$. In other words, there is a symmetry between partitions with respect to probability of random drawing of a larger observation. This indicates the protected attribute has no influence on targets and is independent on outcome. Under the alternative hypothesis, however, the probability of an observation from partition $A$ exceeding an observation from partition $B$ is not equal to $0.5$.

The test involves the calculation of a statistic, usually called \textit{U}, whose distribution under the null hypothesis is known. To calculate Mann Whitney \textit{U}, one first ranks all the values from low to high, paying no attention to which group each value belongs. The smallest number gets a rank of 1. The largest number gets a rank of $|\mathcal{D}|$, where $|\mathcal{D}|=|\mathcal{D}_A|+|\mathcal{D}_B|$ is the total number of values in the two partitions.
\begin{definition}[Mann Whitney \textit{U}]
Given binary partitions $A$ and $B$, the the Mann Whitney \textit{U} statistic on $A$ is defined as,
\begin{align}
\label{eq:U}
    U = r_A - \frac{|\mathcal{D}_A|(|\mathcal{D}_A|+1)}{2}
\end{align}
where $|\mathcal{D}_A|$ is the sample size of $\mathcal{D}_A$, $r_A = \sum_{i=1}^{|\mathcal{D}_A|}rank(y_i)_A$ denotes the sum ranks in partition $A$. The rank of an observation in a sequence $\{y_1,...,y_n\}$ of an ascending order, denoted as $rank(y_i)$ where $i\in\{1,...n\}$, is its index $i$.
\end{definition}

When computing Mann Whitney \textit{U}, the number of comparisons equals the product of the number of values in partition $A$ times the number of values in partition $B$, i.e. $U = \alpha\times|\mathcal{D}_A|\times|\mathcal{D}_B|$, where $\alpha\in(0,1)$. If the null hypothesis is true, then the value of \textit{U} should be half that value, i.e. $\alpha=0.5$. Thus, under the strict fairness, the Mann Whitney \textit{U}, $U_{fairness}$, satisfies
\begin{align}
\label{eq:U_fairness}
    U_{fairness}=0.5\times|\mathcal{D}_A|\times|\mathcal{D}_B|
\end{align}
Note that Mann Whitney \textit{U} test is able to be extended to multi-class $Z$, due to space limit, in this paper, we only focus on binary protected variable.

\section{Problem Formulation}
We consider a dataset $\mathcal{D} = \{(\mathbf{x}_i^j,y_i^j)\}_{i=1}^{h}, j=1,...,k, i=1,...h$, where $\mathbf{x}_i^j\in\mathbb{R}^n$ denotes the $i$-th observation for the $j$-th task, $y_i^j$ denotes the corresponding output and $h$ is the number of observations for each task. Let $\mathit{X}^j = [\mathbf{x}_1^j,...,\mathbf{x}_{h}^j]^T\in\mathbb{R}^{{h}\times n}$ denote the data matrix which contains $n$ data attributes, including binary protected attributes $\mathbf{z}^j = [z_1^j,...,z_h^j]^T\in\mathbb{R}^h$, where $z_i^j \in\{0,1\}$ and explanatory attributes. $\mathbf{y}^j = [y_1^j,...,y_{h}^j]^T\in\mathbb{R}^{h}$ denotes output for $j$-th task. The weight vector for all $k$ tasks from the weight matrix $W=[\mathbf{w}^1,...,\mathbf{w}^k]^T\in\mathbb{R}^{k\times n}$, where $\mathbf{w}^j\in\mathbb{R}^n$ is the weight vector for the $j$-th task, needs to be estimated from the data. The problem, hence, is formulated as:
\noindent\fbox{%
    \parbox{8.7cm}{%
\textbf{Problem: }Fairness Learning in A Multi-task Regression Problem with $k$ tasks.\\
\textbf{Given: }
\begin{itemize}
    \item Data $\mathcal{D} = \{(\mathbf{x}_i^j,y_i^j)\}_{i=1}^{h},\quad j=1,...,k,\quad i=1,...,h$ and data size of two partitions, $|\mathcal{D}_A|$ and $|\mathcal{D}_B|$
    \item $\mathbf{z}^j = [z_1^j,...,z_h^j]^T\in\mathbb{R}^h,\quad j=1,...k$, where $z_i^j \in\{0,1\}$, the binary protected attribute, is included in $X^j=[\mathbf{x}_1^j,...,\mathbf{x}_{h}^j]^T\in\mathbb{R}^{{h}\times n}$.
    \item $\mathbf{p} = [p_1,...,p_{h\times j}]^T\in\mathbb{R}^{h\times j}$, where $p_i=rank(y_i)$, a vector of ranks over all instances
\end{itemize}
\textbf{Goal: }Weight matrix $W=[\mathbf{w}^1,...,\mathbf{w}^k]^T\in\mathbb{R}^{k\times n}$, where $\mathbf{w}^j\in\mathbb{R}^n$ is the weight vector for the $j$-th task, learned from the regression model is bias-free. Thus, predictive outputs $\hat{Y}=\{\hat{y}_i^j\}_{i=1}^h, \quad j=1,...,k, \quad i=1,...,h$, is independent on the protected attribute $Z$.
    }%
}

Therefore, we propose to use multi-task learning model for regression which  was stated in \cite{Liu2009} and penalized with $\ell_{2,1}$-norm, enhanced with the fairness constraint (i.e. the independence test from the previous section), to jointly learn group classifier that leverage information between tasks. The MTL loss function $L(W)$ becomes
\begin{align}
    \min_W \quad &L(W)=\frac{1}{2} \sum_{j=1}^k ||X^j\mathbf{w}^j-\mathbf{y}^j||_F^2 + \beta||W||_{2,1}  \label{eq:scorefunction}
\end{align}
where $||W||_{2,1}:=\sum_{j=1}^k||\mathbf{w}^j||_2$ is group regularization and $\beta>0$ is its parameter. Here, in contrast to single-task problem, the $\ell_{2,1}$-norm regularization encourages multiple predictors from various tasks to share similar parameter sparsity patterns. It penalizes the sum of the $\ell_2$-norms of the blocks of coefficients associated with each attribute across tasks, leading to an $\ell_{2,1}$-norm regularized non-smooth convex loss function denoted as $L(W)$.

Strictly speaking, to control the prediction bias in the MTL regression model, we are required to restrict the $U$ score to a fairness level, i.e. $U=U_{fairness}=0.5\times|\mathcal{D}_A|\times|\mathcal{D}_B|$. For the ease of solving the problem, we restricted the constraint to  a  soft  extent,  such  that
\begin{align*}
    0\leq\frac{U}{|\mathcal{D}_A|\times|\mathcal{D}_B|}-0.5\leq \epsilon
\end{align*}
where $\epsilon$ is a user-defined small positive number of threshold. One can calculate a related \textit{U} score (i.e. $U_B$) using partition $B$. $U$ and $U_B$ are complementary in that they always add up to $|\mathcal{D}_A|\times|\mathcal{D}_B|$, similar to the way that flipping two partitions when calculating an \textit{ROC} curve gives you an \textit{AUC} that is one minus the \textit{AUC} of the unflipped curve. In other words, $\frac{U}{|\mathcal{D}_A|\times|\mathcal{D}_B|}$ alternatively equals to \textit{AUC}. Together with the loss function in Eq.(\ref{eq:scorefunction}), now the problem is formulated as :
\begin{align}
\label{eq:probformulation}
    \min_W \quad &L(W)\\
    \text{s.t. } \quad & 0\leq\frac{U}{|\mathcal{D}_A|\times|\mathcal{D}_B|}-0.5\leq \epsilon \nonumber
\end{align}



\section{Problem Transformation}
Furthermore, to simplify the constraint in Eq.(\ref{eq:probformulation}) by taking in Eq.(\ref{eq:U}), we have
\begin{align}
    C \leq r_A \leq C+\kappa
\end{align}
where $C=\frac{|\mathcal{D}_A|(|\mathcal{D}_A|+1)}{2}+\frac{|\mathcal{D}_A|\times|\mathcal{D}_B|}{2}$ and $\kappa=|\mathcal{D}_A|\times|\mathcal{D}_A|\times\epsilon$ are constant only depended on the size of binary partitions (i.e. $|\mathcal{D}_A|$ and $|\mathcal{D}_B|$) and $\epsilon$. \textit{Therefore, independence test of restricting the Mann Whitney \textit{U} value can be transformed to control the sum rank of partition A instead.}

We further reformulate the constraint to a non-convex set $\mathcal{Q}$ defined as,
\begin{align*}
    \mathcal{Q}=\{M\ |\ C\leq r_A(M)\leq C+\kappa,r_A(M)= \\
    \sum_{i=1}^{|\mathcal{D}_A|} rank([X_A W]_i),M=XW\} \nonumber
\end{align*}
where $r_A(M)$ denotes the sum ranks of vectorized $M$ with respect to partition $A$. We refer to the set $\mathcal{Q}$ as a constraint set and assume that the problem is feasible when $\mathcal{Q}$ is nonempty.
\begin{theorem}
\label{theorem:refroemulation}
Assume the objective function $f(x,z):\mathbb{R}^n\times\mathbb{R}^q\rightarrow\mathbb{R}$ is jointly convex in $x$ and $z$. We consider the optimization problem
\begin{align}
\label{eq:theorem1:1}
    \min \; f(x,z) \quad \text{s.t.} \; Ax+Bz=c, \; g(z)\leq d,\; g(z)\geq e
\end{align}
where $x\in\mathbb{R}^n$ and $z\in\mathbb{R}^q$ are decision variables, $A\in\mathbb{R}^{p\times n}, B\in\mathbb{R}^{p\times q}, c\in\mathbb{R}^p$ are problem data, and $d,e\in\mathbb{R}$. Define $\phi:\mathbb{R}^q\rightarrow\mathbb{R}\cup\{-\infty,+\infty\}$ such that $\phi(z)$ is the best objective value of the problem after fixing $z$. Thus, it is equivalent to 
\begin{align}
\label{eq:theorem1:2}
    \min \; \phi(z)=\inf_x\{f(x,z)|Ax+Bz=c\} \quad \text{s.t.} \; z\in\mathcal{Z}
\end{align}
where $\mathcal{Z}\subseteq \mathbb{R}^q$ is closed.
\end{theorem}
\begin{IEEEproof}
Intuitively, the inequality constraints in Eq.(\ref{eq:theorem1:1}) can be converted to a set of $z$ where $z\in\mathcal{Z}$. Therefore, Eq.(\ref{eq:theorem1:1}) is equivalent to 
\begin{align}
\label{eq:prooftheorem1:1}
    \min \; f(x,z) \quad \text{s.t.} \; Ax+Bz=c, \; z\in\mathcal{Z}
\end{align}
Absorbing the linear constraint, finding the lower bound of $\phi(z)$ is equivalent to minimize the problem in Eq.(\ref{eq:prooftheorem1:1}). That is, Eq.(\ref{eq:prooftheorem1:1}) can be further transformed to Eq.(\ref{eq:theorem1:2}).
\end{IEEEproof}

Together with loss function, therefore, the optimization problem is formulated
\begin{align}
    \min_{W,M} \quad &L(W,M) \label{eq:L(W,M)} \\
    \text{s.t. } \quad &M=XW, \quad M\in\mathcal{Q} \nonumber
\end{align}
The loss function here is the same as the $L(W)$ in Eq.(\ref{eq:scorefunction}), except it minimizes over two variables $W$ and $M$. As stated in Theorem \ref{theorem:refroemulation}, we define $\phi$ such that $\phi(M)$ is the best object value of Eq.(\ref{eq:L(W,M)}),
\begin{align*}
    \phi(M) = \inf_{W}\{L(W,M)|XW-M=0\}
\end{align*}
$\phi(M)$ is convex, since it is the partial minimization of a convex function over a convex set.  It is defined over $M$, but we are interested in finding its minimum value over the non-convex set $\mathcal{Q}$. In other words, the problem can be finally reformulated in a form of minimizing a convex score function over a non-convex set,
\begin{align}
    \min_M \quad &\phi(M) \\
    \text{s.t. } \quad &M\in\mathcal{Q} \nonumber
\end{align}

\section{NC-ADMM Framework}
In this section, we describe the non-convex alternating direction method of multipliers (NC-ADMM) \cite{Diamond2017} as a mechanism to carry out local search methods to solve the optimization problem by returning a local solution. The goal of NC-ADMM is to apply ADMM as a heuristic to solve nonconvex problems and has been explored \cite{Derbinsky2013} as a message passing algorithm. This technique is a new framework that has promising results in a number of applications by minimizing convex functions over non-convex sets \cite{Diamond2017,Takapoui2017}. Applying NC-ADMM, we first captured the primal variable with a closed form in the convex proximal step of NC-ADMM by using soft thresholding formula. Furthermore, the dual variable was approximated with two designed projected algorithms. Both of these two algorithms were intended to adjust the sum rank of the protected attribute to satisfy the non-convex constraint.

Solving the problem using NC-ADMM, it has the form,
\begin{align}
    M^{t+1} &:=\arg\min_M(\phi(M)+\frac{\rho}{2}||M-M_S^t+V^t||_2^2) \label{eq:ncadmm1}\\
    M_S^{t+1} &:=\prod_\mathcal{Q} (M^{t+1}+V^t) \label{eq:ncadmm2}\\
    V^{t+1} &:= V^t+M^{t+1}-M_S^{t+1} \label{eq:ncdamm3}
\end{align}
where $\rho>0$ is an algorithm parameter, $t$ is the iteration counter, $\prod_\mathcal{Q}$ is the projection onto $\mathcal{Q}$, $M_S$ is the projected variable for $M$ onto set $\mathcal{Q}$, and $V$ is the scaled dual variable. Note that the $M$-minimization step (i.e. Eq.(\ref{eq:ncadmm1})) is convex since $\phi$ is convex, but the $M_S$-update (i.e. Eq.(\ref{eq:ncadmm2})) is projection onto a non-convex set.

\subsection{Convex Proximal Step.}
The first step (Eq.(\ref{eq:ncadmm1})) of the NC-ADMM algorithm involves solving the convex optimization problem
\begin{align*}
    \min_{W,M} \quad &\frac{1}{2}||XW-Y||_F^2+\beta||W||_{2,1}+\frac{\rho}{2}||M-M_S^t+V^t||_2^2 \\
    \text{s.t. }\quad &XW-M=0\nonumber
\end{align*}
This is the original problem (Eq.(\ref{eq:L(W,M)})) with non-convex constraint $\mathcal{Q}$ removed, and an additional convex quadratic term involving $M$ added. Note that the problem above has two variables $W$ and $M$ and the objective function is separable in the form of $f(W)+g(M)$. Alternating direction method (ADM) stated in \cite{Deng2013}, thus is applicable. 
The augmented lagrangian problem is of the form
\begin{align*}
    \min_{W,M} \quad &\frac{1}{2}||X W-Y||^2_F+\beta||\mathit{W}||_{2,1}+\frac{\rho}{2}||M-M_S^t+V^t||_2^2 \\
    &- \mathbf{\lambda}^T(XW-M) + \frac{\gamma}{2} ||XW-M||_2^2 \nonumber
\end{align*}
where $\mathbf{\lambda}$ is multiplier and $\gamma$ is penalty parameter. 
This has a closed form solution by soft thresholding formula:
\begin{align*}
    W^j = \max \Bigg\{||c^j||_2-\frac{\beta}{\gamma+1},0 \Bigg\}\frac{c^j}{||c^j||_2}, \quad \text{for } j=1,...,k
\end{align*}
where $c^j:= (X^j)^+\cdot\frac{Y^j+\gamma M^j+\lambda^j}{1+\gamma}$ and $(X^j)^+$ denotes the pseudo-inverse of $X$ for the $j$-th task. We denote $W = group((X)^+\cdot\frac{Y+\gamma M+\lambda}{1+\gamma})$ as group-wise shrinkage for short. $\lambda$ is updated with step size $\theta$ $(\theta>0)$. The output $M^{t+1}$ and $W^{t+1}$ from the convex proximal step of NC-ADMM framework thus were passed on to projection of Eq.(\ref{eq:ncadmm2}).

\subsection{Projection onto Non-Convex Set.} 
The non-convex projection step consists of finding a closest point, $M_S^{k+1}$, in $\mathcal{Q}$ to $M^{k+1}+V^k$. Suppose that a feasible solution $M_S^{k+1}$ can be reached, the projection is equivalent to maximize the sum of square of the non-zero elements in $M_S^{k+1}$, as stated in our Theorem \ref{theorem:projection}.

\begin{theorem}\label{theorem:projection}
Let $\mathbf{b}, \mathbf{v}$ and $\mathbf{b}_S$ are vectorized $M,V$ and $M_S$ respectively and $\mathbf{b}_S$ is the projected vector of $\mathbf{b+v}$ onto the non-convex set $\mathcal{Q}$. Then the projection 
\begin{align}
    M_S^{t+1} &:= \prod_\mathbf{Q} (M^{t+1}+V^t) \nonumber\\
              &= \arg\min_{M_S^{t+1}\in\mathcal{Q}}\sum_{i=1}^{k\times h} ([\mathbf{b}_S^{t+1}]_i-[\mathbf{b}^{t+1}+\mathbf{v}^t]_i)^2
\end{align}
is equivalent to below 
\begin{align}
    \arg\max_{S\subseteq\{1,...,k\times h\},M_S^{t+1}\in\mathcal{Q}}\sum_{i\in S}[\mathbf{b}_S^{t+1}]_i^2 \label{eq:projection2}
\end{align}
where $[\mathbf{b}^{t+1}_S]_i$ indicates the $i$-th element in $\mathbf{b}^{t+1}$ and $S$ is the set of indices which predicted targets are non-zero.
\end{theorem}

\begin{IEEEproof}
We denote $S_A$ and $S_B$ as sets of indices of observations in $\mathcal{D}$ for binary partition A and B, respectively, with $S_A\cup S_B = \{1,...,k\times h\}$ and define $\mathbf{d}^{t+1}:=\mathbf{b}^{t+1}+\mathbf{v}^t$. Since $S\subseteq(S_A\cup S_B)$ and $\mathbf{d}^{t+1}$ is a vector with ranks, it is easy to find that $\mathbf{d}^{t+1}-\mathbf{b}_S^{t+1}$ is non-negative, because $\mathbf{b}_S^{t+1}\in\mathcal{Q}$ and $[\mathbf{b}_S^{t+1}]_i=[\mathbf{d}^{t+1}]_i$, for $i\in S$; $[\mathbf{b}_S^{t+1}]_i=0$, otherwise. Therefore, to minimize $||\mathbf{d}^{t+1}-\mathbf{b}_S^{t+1}||_2^2$ is equivalent to minimize $-||\mathbf{b}_S^{t+1}||_2^2$ and further equivalent to maximize $||\mathbf{b}_S^{t+1}||_2^2$, all with constraint $M_S^{t+1}\in\mathcal{Q}$.
\end{IEEEproof}
From the previous section, we introduced constraint $\mathcal{Q}$ by limiting $r_A$ to a predefined level $[C,C+\kappa]$ which is only dependent on $|\mathcal{D}_A|$ and $|\mathcal{D}_B|$. We, hence, proposed two approximate projection algorithms and both follow the rules:
\begin{itemize}
    \item Adjust (increase or decrease) the sum rank of partition A of predictive data, $r_A$, falls between the restricted level $[C,C+\kappa]$ by removing some predictive targets $\Hat{y}_i$ to 0, where $i\in S', S':= (S_A\cup S_B)\backslash S$. This guarantees the Mann Whitney \textit{U} value is close enough to $U_{fairness}$, such that the predicted target is independent on the protected attribute.
    \item Summation of non-zero value of the predicted targets (i.e. $\sum_{i\in S}\Hat{y}_i$) is maximum.
\end{itemize}
According to the projection rules we proposed above, we then divided the projection into four distinct scenarios. Key steps are described in Algorithm \ref{alg:ProjectionToQ}.

\vspace{-3mm}
\begin{algorithm}[h]
\caption{A Projection Algorithm onto $\mathcal{Q}$.}
\textbf{Input: }$M$, $M_S$ and $V$.\\
\textbf{Initialization: }Input were initialized with random samples from a uniform distribution over $[0,1)$.\\
\textbf{Output: }$M_S$ onto the constraint set $\mathcal{Q}$.
\begin{algorithmic}[1]
\State Update data ranks with $M_P:=M+V$;
\State Calculate $C, \kappa, r_A$ and $r_A^{most}$;
\If{$r_A^{most}<C$} 
    \State No feasible solution is obtained;
\ElsIf{$C\leq r_A\leq C+\kappa$}
    \State $M_S = M_P$;
\ElsIf{$C+\kappa<r_A$}
    \State Apply projected Algorithm \ref{alg:Projection1};
\ElsIf{$r_A<C$}
    \State Apply projected Algorithm \ref{alg:Projection2};
\EndIf
\State \Return $M_S$
\end{algorithmic}
\label{alg:ProjectionToQ}
\end{algorithm}
\setlength{\textfloatsep}{0pt}
\vspace{-3mm}

$r_A^{most}<C$. We denote $r_A^{most}$ as the sum rank of partition $A$ when the smallest rank in $A$ is larger than the largest rank in $B$. This scenario indicates both projection algorithms fail. Hence, no feasible solution is obtained.

$C\leq r_A \leq C+\kappa$. Constraint is satisfied and no projection is needed.

$C+\kappa<r_A$. This indicates the sum rank of $A$ went beyond the upper limit of the constraint. Therefore, in order to decrease $r_A$ and make it falls in the constraint, some targets in $A$ are required to be 0. Algorithm \ref{alg:Projection1} was developed. Thus an approximated result was obtained. In step 2, $d>0$ indicates the rank difference required to be cut off by $r_A$. We denote $\Delta$ as the the absolute difference of $r_A$ when $S'=S_A$. Therefore the problem can be seen to find a subset of $S_A$, i.e. $S'$, such that $r_A$ satisfies the constraint. Apparently, traversing all subsets is expensive. 
We assume every subset of $S_A$ has a integer index number which can be converted to a corresponding binary number and further find its corresponding subset. 
For example, $S_A$ has a cardinality of 4. The index of its empty subset is 0 and the corresponding binary is $\{0,0,0,0\}$. In addition, the index and the corresponding binary of its universe set are 15 and $\{1,1,1,1\}$, respectively.
$j$, hence, is the corresponding index of approximate $S'$. Since repeated ranks may occur due to the same target value, a reasonable parameter of searching range $p$ was applied to collect all candidate indices $\mathcal{J}$. In step 4, the optimal index $m$ was selected where the difference between $\Delta_m$ and $d$ is minimum.

\begin{algorithm}[h]
\caption{An Approximate Projected Algorithm When $C+\kappa<r_A$.}
\textbf{Input: }$M$ conditional on $C+\kappa<r_A$.\\
\textbf{Output: }$M_S$ onto the constraint set $\mathcal{Q}$.
\begin{algorithmic}[1]
\State Initialize searching range $p$;
\State $d=|C-r_A|$; 
\State $j=2^{|S_A|}\times \frac{d}{\Delta}, \mathcal{J}=\{ \floor*{j}-p,..., \floor*{j}, \ceil*{j},...,\ceil*{j}+p\}$;
\State $m=\arg_{m\in\mathcal{J}}\min |\Delta_m-d|$;
\State Find the corresponding $S'$ with $m$;
\State (Optional: check $r_A$ and apply Algorithm \ref{alg:Projection2}) if $r_A<C$;
\State $S = (S_A\cup S_B)\backslash S'$ ;
\State \Return $M_S$ 
\end{algorithmic}
\label{alg:Projection1}
\end{algorithm}
\setlength{\textfloatsep}{0pt}
\vspace{-3mm}

$r_A<C$. It is similar to the first scenario but increasing $r_A$ by removing some target values in partition $B$ to 0. An approximate Algorithm \ref{alg:Projection2} is given. First of all, for each instance $i$ in partition $B$, calculating the number of instances $m_i$ of $A$ which is lower than $i$'s rank. This is presented in step 2 to 4. We further applied a greedy algorithm to add $B$ instance to $S'$ in an ascending rank order. The greedy algorithm stops until $r_A$ satisfied the constraint.

\vspace{-2mm}
\begin{algorithm}[h]
\caption{An Approximate Projected Algorithm When $r_A<C$.}
\textbf{Input: }$M$ conditional on $r_A<C$.\\
\textbf{Output: }$M_S$ onto the constraint set $\mathcal{Q}$.
\begin{algorithmic}[1]
\State $d=|C-r_A|, \quad l=0$
\For{each data $i$ where $z_i=B$}
    \State $m_i=\sum_{j\in S_A} I_{\{rank_i>rank_j\}}$
\EndFor 
\While{$l\notin [d,C+\epsilon-r_A]$}
    \State $l=l+m_i$
    \State $i$ is added to $S'$
\EndWhile
\State $S=(S_A\cup S_B)\backslash S'$
\State \Return $M_S$ 
\end{algorithmic}
\label{alg:Projection2}
\end{algorithm}
\setlength{\textfloatsep}{0pt}
\vspace{-5mm}

\subsection{Dual Update.}
The last part of the NC-ADMM framework is to update the scaled dual variable $V^t$, which is able to be interpreted as the running sum of error of $M^{t+1}-M_S^{t+1}$.

\subsection{Framework Analysis.}
Firstly, since the convex proximal step was solved by ADM which computes the exact solution for each subproblem, its convergence is guaranteed by the existing ADM theory \cite{Glowinski1989} and costs $O(p\cdot k\cdot h^3)$, where $p$ is the number of iterations and we assumed $k>h$. Complexity of the projection Algorithm \ref{alg:Projection1} is fully dependent on the finding range $q$ and takes $O(q\cdot\log q)$ running time and Algorithm \ref{alg:Projection2} takes $O(k^2\cdot h^2)$. Therefore, together with the NC-ADMM framework, the total complexity is $O(s\cdot (q\log q+k^2h^2))$, where $s$ is the iterated number of NC-ADMM. When the projected set $\mathcal{Q}$ is convex, ADMM is guaranteed to converge to a solution with a global optima. However, in the proposed problem, $\mathcal{Q}$ is non-convex and NC-ADMM fails to guarantee convergence or a global minimum \cite{Diamond2017}. For this reason, in the following section, $s$ was set with a fixed number for each experiment.

\section{Experimental Settings}
\subsection{Evaluation Metrics.}
To evaluate the proposed techniques for fairness learning, we introduced several classic evaluation metrics to measure data biases.
These measurements came into play that allows quantifying the extent of bias taking into account the protected attribute and were designed for indicating indirect discrimination.

\subsubsection{The area under the ROC curve (AUC)} 
$= \frac{\sum_{(z_i,y_i)\in\mathcal{D}_A}\sum_{(z_j,y_j)\in\mathcal{D}_B}I(rank(y_i)>rank(y_j))}{|\mathcal{D}_A|\times|\mathcal{D}_B|}$
where $I(\cdot)$ is an indicator function which returns 1 if its argument is true, 0 otherwise and $rank(y)$ is the rank of $y$ over entire data. $AUC=0.5$ represents random predictability, thus $Z$ is independent on $Y$. 


\subsubsection{Mean Difference (MD)\cite{Calders2013,Kamiran2010}}
$= \frac{\sum_{(z_i,y_i)\in\mathcal{D_A}}y_i}{|\mathcal{D}_A|} - \frac{\sum_{(z_i,y_i)\in\mathcal{D_B}}y_i}{|\mathcal{D}_B|}$
It measures the difference between the means of the continuous targets $y$ over the protected attributes $z$ of data $\mathcal{D}$, partitioned into $\mathcal{D}_A$ and $\mathcal{D}_B$. If there is no difference (i.e. $MD=0$) then it is considered that there is no discrimination which indicates the independent relation of the protected attribute and targets.


\subsubsection{Balanced Residuals (BR)\cite{Calders2013}}
$= \frac{\sum_{i\in\mathcal{D}_A} y_i-\Hat{y}_i}{|\mathcal{D}_A|} - \frac{\sum_{i\in\mathcal{D}_B} y_i-\Hat{y}_i}{|\mathcal{D}_B|}$
It characterizes the difference between the actual outcomes $y$ and the model outputs $\Hat{y}$ of data $\mathcal{D}$, partitioned into $\mathcal{D}_A$ and $\mathcal{D}_B$, respectively. This requirement is that underpredictions and overpredictions should be balanced within the protected attribute. A value of zero signifies no dependency of the protected attribute and targets. Positive $BR$ would indicate the bias towards partition A.


\subsubsection{Impact Rank Ratio (IRR)\cite{Pedreschi2009}}
$= \frac{r_A}{|\mathcal{D}_A|}\Big/\frac{r_B}{|\mathcal{D}_B|}$
It is defined as the ratio of mean sum rank of protected partition $A$ over it of partition $B$ in data $\mathcal{D}$. The decisions are deemed to be discriminatory if the ratio of positive outcomes for the protected attribute is below $80\%$. $IRR=1$ indicates that there is no bias of data $\mathcal{D}$.


\begin{table*}[h]
\small
\centering
\captionof{table}{Consolidated Overall Result For Synthetic Data} 
\label{tab:syntable}
\begin{tabular}{ |c|c|c|c|c|c|c|c|c|c|c|c| } 
\hline
& \multirow{2}{*}{Data} & \multirow{2}{*}{Our Method (Rank)} & \multicolumn{4}{c|}{SSEM} & \multicolumn{4}{c|}{SSBR} & \multirow{2}{*}{SDBC} \\\cline{4-11}
& & & s=2 & s=3 & s=4 & s=5 & s=2 & s=3 & s=4 & s=5 &  \\
\hline
\multicolumn{12}{|c|}{Synthetic data, $\alpha$ = 0.9}\\
\hline
AUC & 0.902 & $\mathbf{0.505\pm0.004}$ & 0.558 & 0.525 & 0.547 & 0.530 & 1.000 & 0.985 & 0.974 & 0.964 & $0.528\pm0.030$\\
\hline
MD & -0.187 & $-0.080\pm0.019$ & 0.224 &0.044&0.114&$\mathbf{0.024}$&-0.061&-0.069&-0.188&-0.257&$0.497\pm0.003$\\ 
\hline
BR & - &$-0.273\pm0.020$&-0.358&-0.272&-0.361&-0.325&$\mathbf{-0.211}$&-0.275&-0.248&-0.229&$-0.742\pm0.109$\\
\hline
IRR & 0.458&$\mathbf{0.991\pm0.008}$&0.892&0.952&0.912&0.942&0.375&0.387&0.396&0.404&$0.947\pm0.058$\\
\hline
\multicolumn{12}{|c|}{Synthetic data, $\alpha$ = 0.8}\\
\hline
AUC & 0.807&$\mathbf{0.501\pm0.002}$&0.552&0.531&0.531&0.525&0.946&0.887&0.883&0.873&$0.534\pm0.072$\\
\hline
MD &-0.407 &$-0.038\pm0.015$ & 0.146&0.134&0.207&0.064&$\mathbf{-0.027}$&0.058&-0.119&-0.220&$0.205\pm0.017$\\
\hline
BR & - &$-0.341\pm0.132$&-0.452&-0.445&-0.596&-0.459&-0.315&-0.350&-0.345&$\mathbf{-0.267}$&$-0.594\pm0.014$\\
\hline
IRR & 0.553 &$\mathbf{0.998\pm0.004}$&0.902&0.941&0.941&0.953&0.420&0.473&0.476&0.485&$0.945\pm0.131$\\
\hline
\multicolumn{12}{|c|}{Synthetic data, $\alpha$ = 0.7}\\
\hline
AUC &0.708 & $\mathbf{0.508\pm0.003}$&0.575&0.532&0.534&0.522&0.859&0.809&0.783&0.767&$0.524\pm0.080$\\
\hline
MD &0.272&$\mathbf{0.010\pm0.005}$&0.129&0.032&0.176&0.119&-0.178&-0.093&-0.171&-0.193&$0.102\pm0.027$\\
\hline
BR &-&$\mathbf{-0.262\pm0.005}$ &-0.417&-0.349&-0.515&-0.469&-0.264&-0.324&-0.290&-0.289&$0.459\pm0.023$\\
\hline
IRR & 0.669&$\mathbf{0.984\pm0.006}$ &0.863&0.940&0.936&0.958&0.499&0.551&0.578&0.597&$0.861\pm0.100$\\
\hline
\multicolumn{12}{|c|}{Synthetic data, $\alpha$ = 0.6}\\
\hline
AUC &0.608&$0.508\pm0.013$&$\mathbf{0.497}$&0.541&0.528&0.566&0.805&0.758&0.781&0.778&$0.575\pm0.076$\\
\hline
MD &-0.254&$\mathbf{0.027\pm0.054}$&-0.065&-0.035&0.052&0.153&-0.083&-0.420&-0.227&-0.042&$-0.140\pm0.449$\\
\hline
BR &-&$\mathbf{-0.075\pm0.161}$&-0.340&-0.387&-0.450&-0.536&-0.420&-0.256&-0.330&-0.445&$-0.346\pm0.132$\\
\hline
IRR &0.709&$0.986\pm0.024$&$\mathbf{1.006}$&0.922&0.947&0.878&0.547&0.601&0.574&0.578&$0.820\pm0.141$\\
\hline
\end{tabular}
\raggedright	
\\\textit{Note: Best performance are labeled in bold.}
\vspace{-3mm}
\end{table*}

\begin{table*}[h]
\small
\centering
\captionof{table}{Consolidated Overall Result For Real Data} 
\label{tab:realtable}
\begin{tabular}{ |c|c|c|c|c|c|c|c|c|c|c|c| } 
\hline
& \multirow{2}{*}{Data} & \multirow{2}{*}{Our Method (Rank)} & \multicolumn{4}{c|}{SSEM} & \multicolumn{4}{c|}{SSBR} & \multirow{2}{*}{SDBC} \\\cline{4-11}
& & & s=2 & s=3 & s=4 & s=5 & s=2 & s=3 & s=4 & s=5 &  \\
\hline

\multicolumn{12}{|c|}{Crime Data}\\
\hline
AUC & 0.728 & $\mathbf{0.495\pm0.001}$ & 0.686 & 0.778 & 0.781 & 0.799 & 0.815 & 0.814 & 0.815 & 0.809 & $0.680\pm0.006$\\
\hline
MD & 0.100 & $\mathbf{-0.010\pm0.009}$ & 0.176 &0.193 &0.198&0.210&0.179&0.185&0.184&0.208&$-0.705\pm0.015$\\ 
\hline
BR & - &$0.111\pm0.009$&$\mathbf{-0.075}$&-0.093&-0.097&-0.109&-0.079&-0.085&-0.084&-0.107&$0.214\pm0.013$\\
\hline
IRR & 0.619 &$\mathbf{1.007\pm0.001}$&0.703&0.595&0.592&0.574&0.558&0.558&0.558&0.563&$0.711\pm0.008$\\
\hline

\multicolumn{12}{|c|}{Income Data}\\
\hline
AUC & 0.815&$\mathbf{0.489\pm0.010}$&0.739&0.771&0.680&0.788&0.988&0.987&0.835&0.986&$0.610\pm0.074$\\
\hline
MD &0.031 &$\mathbf{-0.003\pm0.005}$ & 0.387&0.303&-0.697&0.221&0.412&0.325&-0.698&0.233&$-0.083\pm0.484$\\
\hline
BR & - &$\mathbf{-0.014\pm0.084}$&-0.320&-0.249&-0.028&-0.172&-0.345&-0.369&-0.027&-0.196&$-0.164\pm0.264$\\
\hline
IRR & 0.567 &$\mathbf{1.022\pm0.020}$&0.646&0.611&0.715&0.594&0.429&0.430&0.549&0.430&$0.820\pm0.114$\\
\hline

\multicolumn{12}{|c|}{Wine Data}\\
\hline
AUC & 0.769&$\mathbf{0.489\pm0.014}$&0.764&0.913&0.984&0.925&0.998&0.997&0.995&0.994&$0.777\pm0.026$\\
\hline
MD &-0.563 &$\mathbf{-0.053\pm0.028}$ & -1.537&-0.530&-0.248&-0.667&-3.062&-2.416&-1.425&-0.884&$-0.775\pm0.137$\\
\hline
BR & - &$-0.351\pm0.217$&-0.356&-0.542&-0.650&-0.513&$\mathbf{-0.280}$&-0.333&-0.445&-0.539&$-0.380\pm0.081$\\
\hline
IRR & 0.611 &$\mathbf{1.020\pm0.029}$&0.622&0.491&0.440&0.482&0.431&0.432&0.433&0.434&$0.610\pm0.016$\\
\hline

\multicolumn{12}{|c|}{Student Data}\\
\hline
AUC & 0.798&$\mathbf{0.507\pm0.015}$&0.559&0.584&0.589&0.587&0.866&0.863&0.862&0.860&$0.483\pm0.036$\\
\hline
MD &-1.949 &$\mathbf{-0.379\pm0.088}$ & -0.845&-0.786&-0.763&-0.783&-1.795&-1.662&-1.643&-1.581&$-0.621\pm0.607$\\
\hline
BR & - &$-1.991\pm0.029$&-1.909&-1.917&-1.912&-1.896&-1.887&-1.888&$\mathbf{-1.881}$&-1.899&$-1.956\pm0.191$\\
\hline
IRR & 0.541 &$0.987\pm0.031$&0.889&0.847&0.839&0.842&0.483&0.485&0.487&0.489&$\mathbf{1.036\pm0.073}$\\
\hline
\end{tabular}
\raggedright	
\\\textit{Note: Best performance are labeled in bold.}
\vspace{-5mm}
\end{table*}

\subsection{Experimental Data}
\textbf{Synthetic Data.} To simulate different degrees of bias on predictive outcomes, we generated four synthetic datasets with various levels of $\alpha$ (0.6, 0.7, 0.8 and 0.9) stated in section III representing distinct degrees of dependence of binary protected attribute on targets. Specifically, for each dataset, we generated 1000 data samples along with binary label (\textit{A} or \textit{B}) uniformly. It was divided into 40 tasks ($k=40$) and each contains 25 observations ($h=25$). Each observation was assigned with a feature vector including the protected attribute and four other explanatory attributes. Targets were generated from two different Gaussian distributions with same standard deviation but different means. Ranks from 1 to 1000 were assigned to each observation for all tasks by an ascending order of target values. If observations shared the same target values, we assigned each observation in a tie its average rank.

\begin{table}
\small
    \centering
    \captionof{table}{Key Characteristics of Real Data}
    \label{tab:key}
    \begin{tabular}{|c|c|c|c|c|}
        \hline
        Data & Crime & Income & Wine & Student\\
        \hline
        $z$ & \multicolumn{2}{|c|}{$\{$Black, non-Black$\}$} &$\{$White, Red$\}$ & $\{$M, F$\}$\\
        \hline
        $y$ & Crime Rate & Income & Alcoholicity & Final Grade\\
        \hline
        $|\mathcal{D}|$ & 41652 & 9426 & 6497 & 1044\\
        \hline
        $|\mathcal{D}_A|$ & 13676 & 2581 & 1599 & 453\\
        \hline
        $|\mathcal{D}_B|$ & 27976 & 6845 & 4898 & 591\\
        \hline
        groups & census tracks & counties & prod. places & birth places \\
        \hline
        $k$ & 801 & 3142 & 89 & 36\\
        \hline
        obs. & weeks & years & prod. years & enroll. time\\
        \hline
        $h$ & 52 & 3 & 73 &29\\
        \hline
        $n$ & 15 & 16 & 11 &31\\
        \hline
    \end{tabular}
\raggedright	
\\\textit{Note: $z$ is the protected attribute; $y$ is the predicted target; $k$ is the number of tasks (groups); for each task, there are $h$ observations; $n$ is the number of features.}
\end{table}

\textbf{Real Data.} We experimented with four real-world datasets: the Chicago Communities and Crime in 2015 (Crime), the Adult Income and Counties in the U.S. (Income), the Wine Quality (Wine) \footnote{https://archive.ics.uci.edu/ml/datasets/wine+quality} and the Student Performance (Student) dataset \footnote{https://archive.ics.uci.edu/ml/datasets/student+performance}. Specifically, the Crime and Income datasets contain information including demographics (e.g. race, gender, age dependency ratio, population, etc.), household, education, unemployment situation, etc. These information were separately collected from \textit{American FactFinder (AFF)}\footnote{https://factfinder.census.gov/faces/nav/jsf/pages/index.xhtml} which is an online and self-service database provided by the U.S. Census Bureau and then aggregated to the final data prepared for experiments. For ease of applying of the multi-task regression model, the Crime data contains 801 census tracks ($k=801$), where each of them is a small geo-location and serves as an unique group. Each census track was furthermore subdivided into 52 time instances ($h=52$), hence, 41652 instances in total. The Income data contains 3142 counties ($k=3142$) over the U.S. by recent three years ($h=3$), 2014 through 2016 and hence there are total 9426 instances. In order to be consistent, each data record in the Wine and Student was randomly assigned with a task and instance and thus they were divided into 89 and 36 tasks as well as 73 and 29 instances for each task, respectively. 
All four datasets were divided into two partitions according to the binary protected variable. According to the data information provided in \textit{UCI repository}, the Wine data contains descriptions of white and red wine, same as the Student data into male and female. In the Crime and Income dataset, we created binary groups: majority ($>70\%$) population of Black and non-Black. Key characteristics for all real data are listed in Table \ref{tab:key}. In the end, since the initial Income data has a relatively fair \textit{AUC} (0.502), a small \textit{MD} (0.002) and a high \textit{IRR} (0.925), for our experiments, we randomly selected some target incomes for non-Black and increased by $75\%$. The \textit{AUC} of this modification became to 0.815 with \textit{MD} value of 0.031 and \textit{IRR} of 0.567. Same reason and data processing were applied to the Wine and Student dataset as referred \cite{Calders2013}.

All the attributes were standardized to zero mean and unit variance for both synthetic data and real-world data and prepared for experiments. 

\begin{figure*}[h]
\centering
    \begin{subfigure}[b]{0.245\textwidth}
        \includegraphics[width=\textwidth]{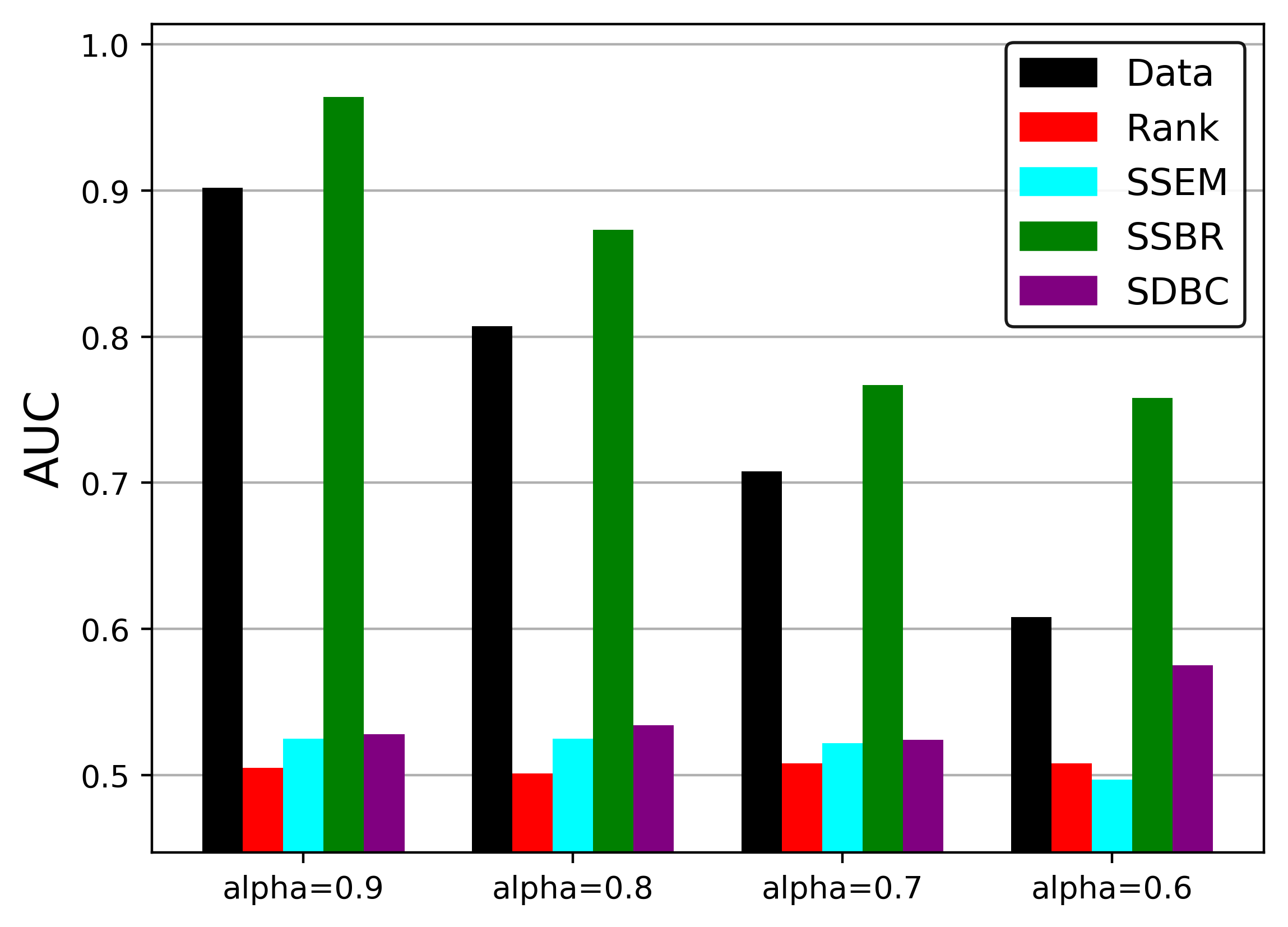}
        \caption{\textit{AUC}, Synthetic Data}
    \end{subfigure}
    \begin{subfigure}[b]{0.245\textwidth}
        \includegraphics[width=\textwidth]{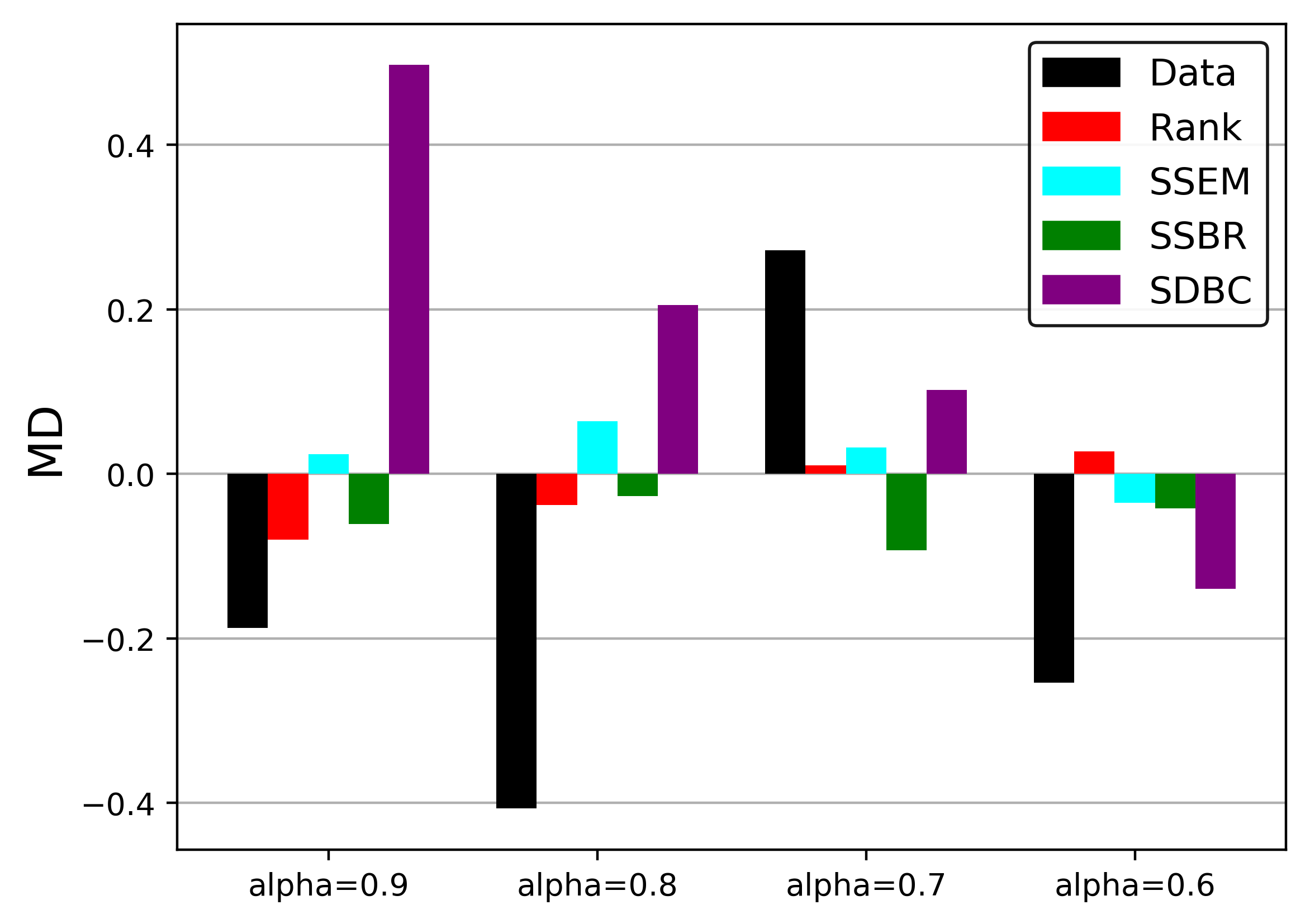}
        \caption{\textit{MD}, Synthetic Data}
    \end{subfigure}
    \begin{subfigure}[b]{0.245\textwidth}
        \includegraphics[width=\textwidth]{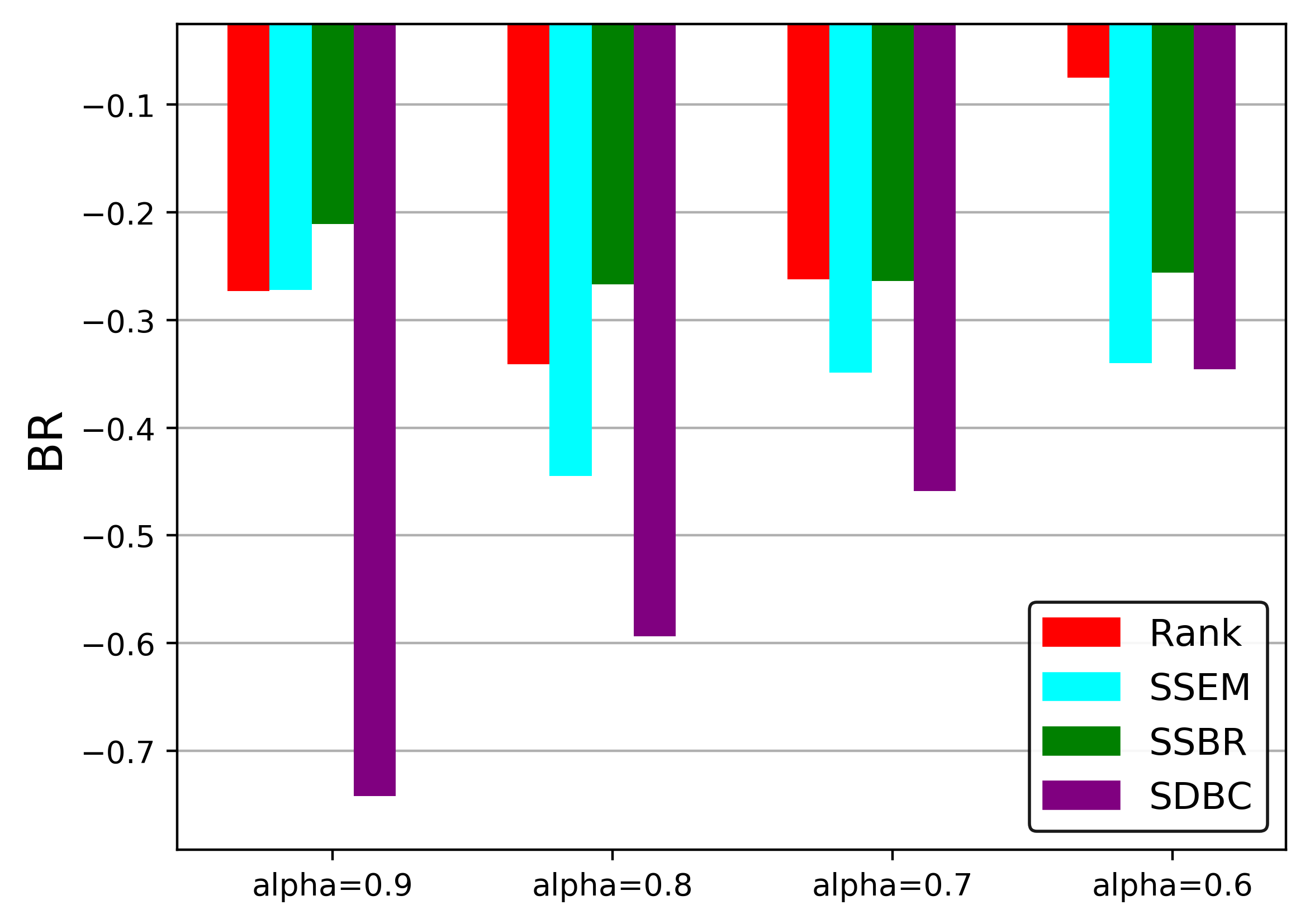}
        \caption{\textit{BR}, Synthetic Data}
    \end{subfigure}
    \begin{subfigure}[b]{0.245\textwidth}
        \includegraphics[width=\textwidth]{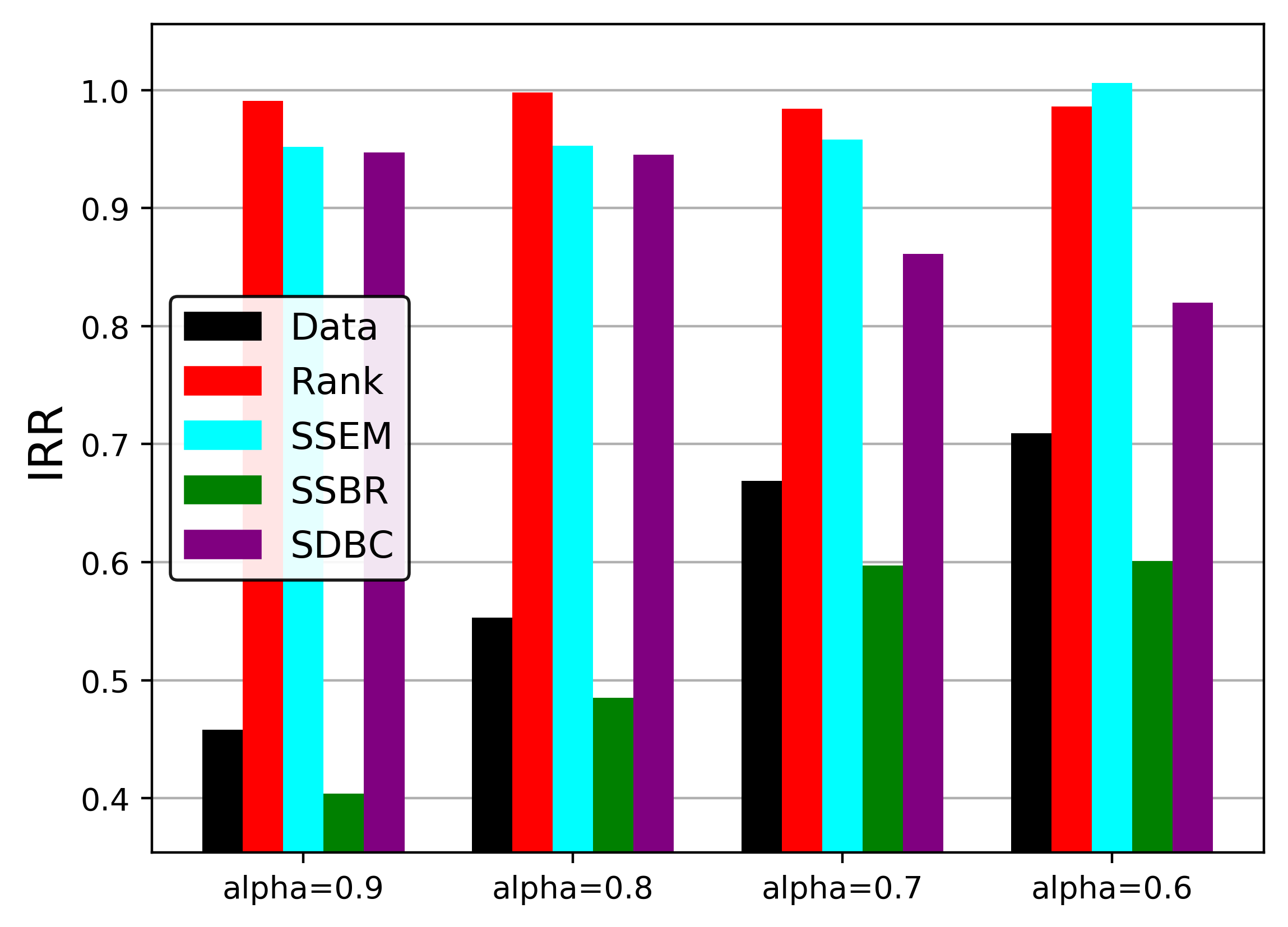}
        \caption{\textit{IRR}, Synthetic Data}
    \end{subfigure}

    \begin{subfigure}[b]{0.245\textwidth}
        \includegraphics[width=\textwidth]{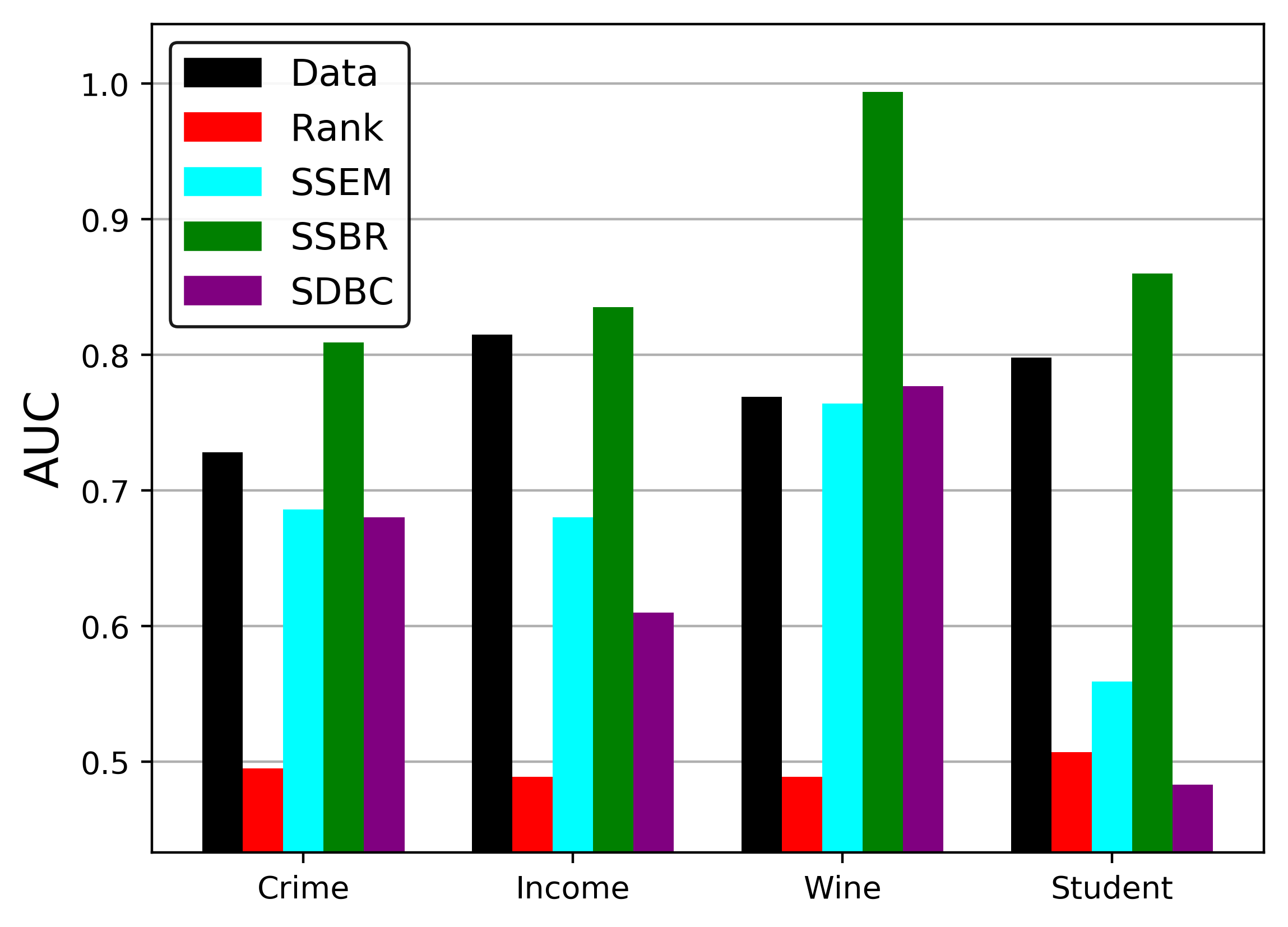}
        \caption{\textit{AUC}, Real Data}
    \end{subfigure}
    \begin{subfigure}[b]{0.245\textwidth}
        \includegraphics[width=\textwidth]{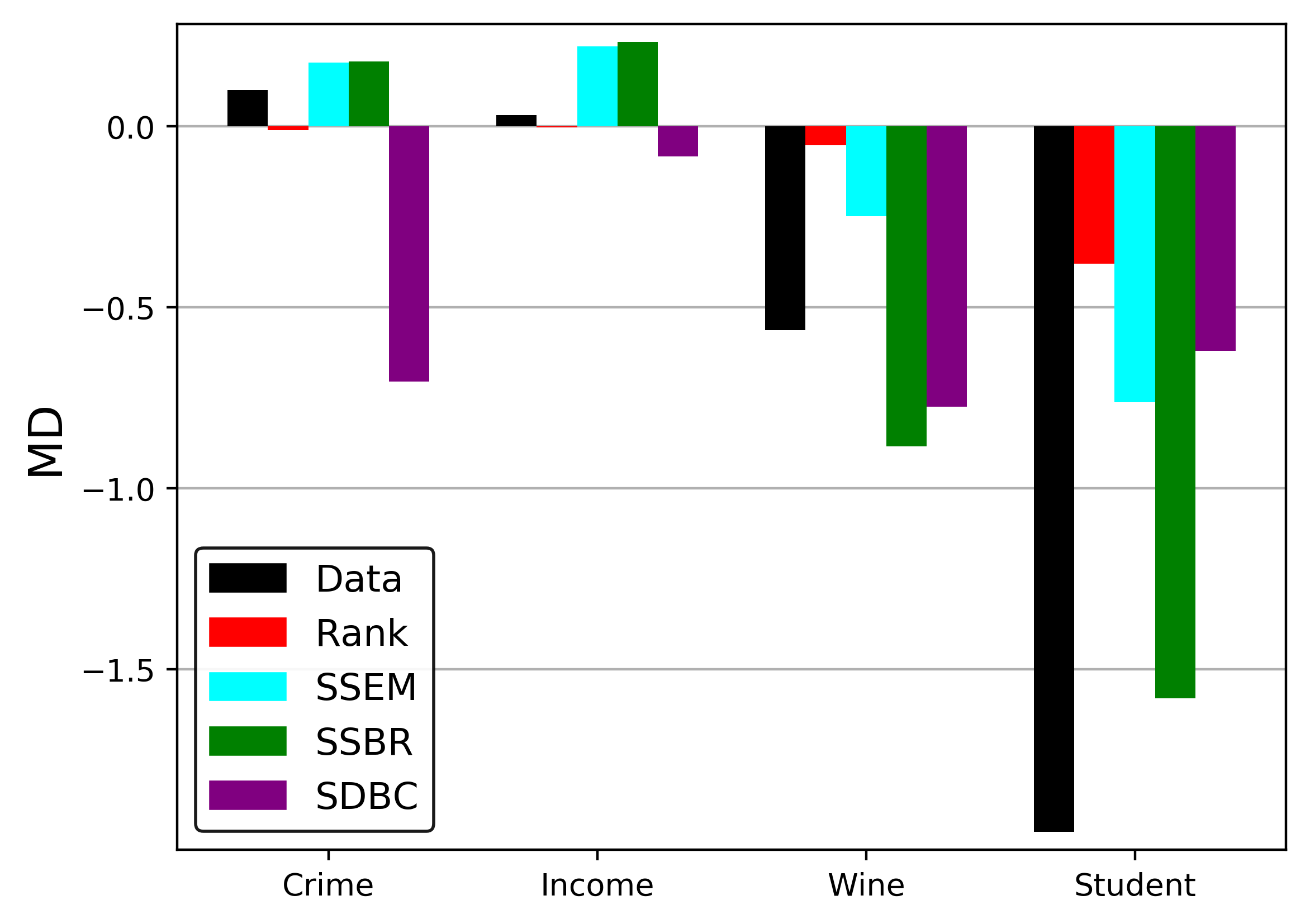}
        \caption{\textit{MD}, Real Data}
    \end{subfigure}
    \begin{subfigure}[b]{0.245\textwidth}
        \includegraphics[width=\textwidth]{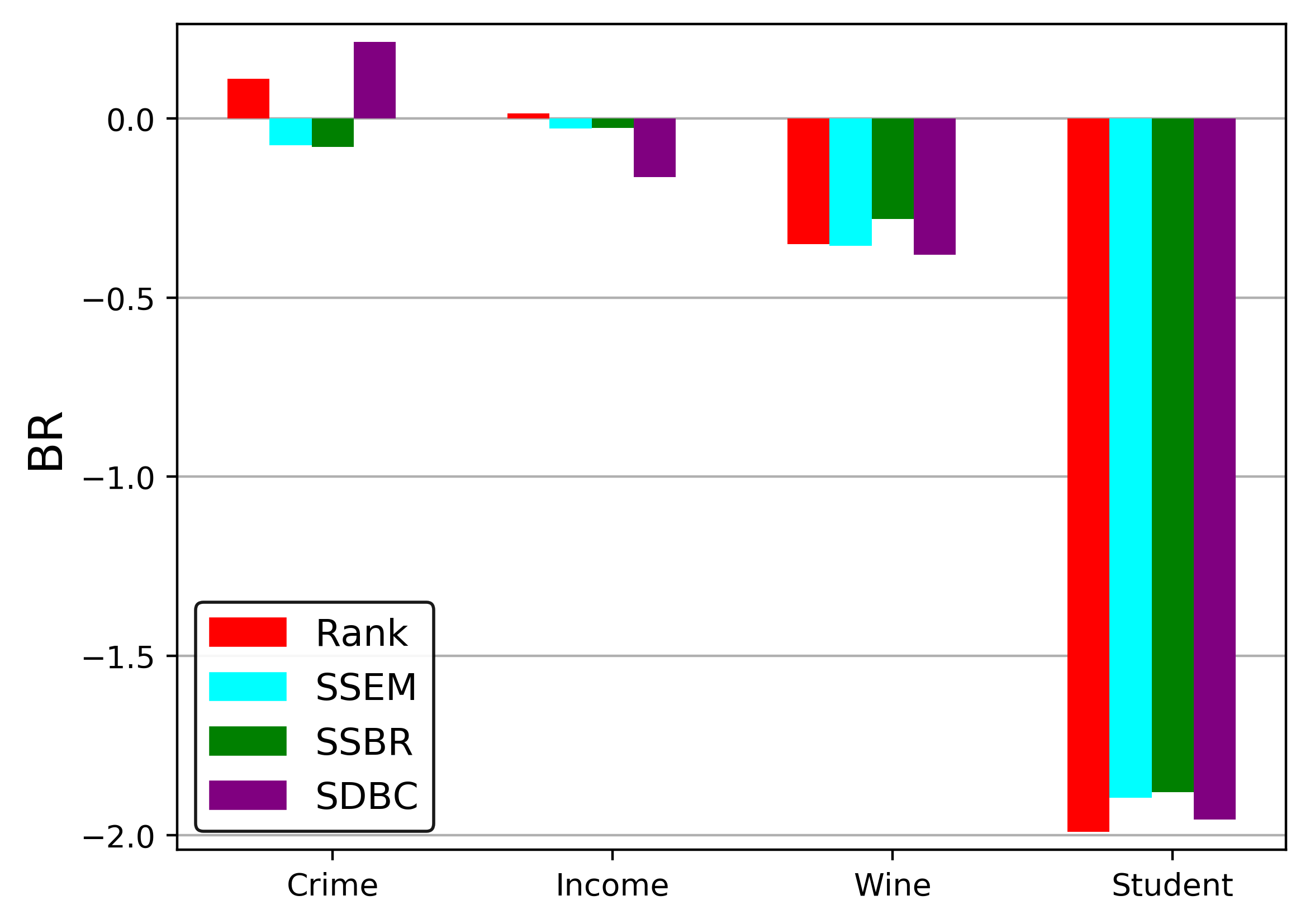}
        \caption{\textit{BR}, Real Data}
    \end{subfigure}
    \begin{subfigure}[b]{0.245\textwidth}
        \includegraphics[width=\textwidth]{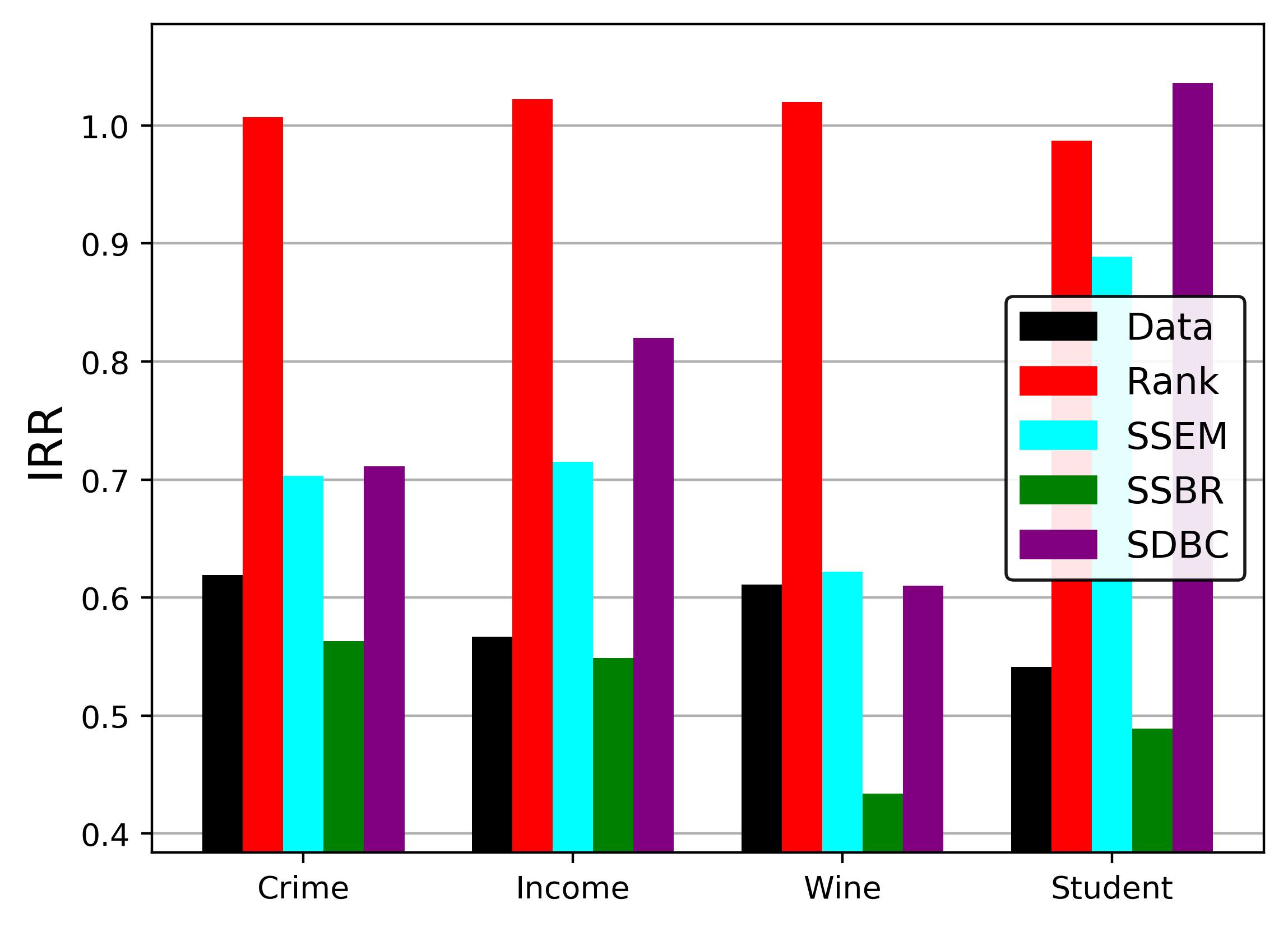}
        \caption{\textit{IRR}, Real Data}
    \end{subfigure}
    \caption{Experiment results for Synthetic and Real data.}
    \label{fig:Syn and Real result}
\vspace{-5mm}
\end{figure*}

\subsection{Baseline Methods.}
\textit{Calders etal.,}\cite{Calders2013} first controlled bias in regression model based on stratification of data with the propensity scoring technique from statistics. 
Two instances having the same propensity score, thus, are expected to receive the same target value. Each stratum was controlled by two convex constraints, separately, i.e. Stratified Strict Equal Means (SSEM) and Stratified  Strict  Balanced  Residual  (SSBR). The goal of SSEM is to learn the coefficient $W$ such that the mean difference between the predictive targets on binary partition \textit{A} and \textit{B} for each stratum is zero. The approach of SSBR is similar to SSEM, except the constraint was replaced with strictly balanced residuals for each stratum. Note that data stratification is dependent on the propensity scores of observations, which disorganizes the structure of multi-tasks. Finally, Strict Decision Boundary Covariance (SDBC) constraint \cite{Zafar2017} was applied as another counterpart methods. Note that the competitive method SDBC is not directly comparable to our proposed methods since it was designed for the classification of a single task. In this work, however, we considered multi-task regression models. SDBC constraint over all tasks indicates the covariance between the protected attribute and predicted target variable is zero. 

\vspace{-2mm}
\subsection{Parameter Tuning.}
For SSEM and SSBR, we set the number of strata to 2, 3, 4 and 5 in order to performance comparison. Baseline method with SDBC was iteratively computed with ADM. Trade-off parameters were selected in a range $[10^{-4}, 10^4]$ and the one returning minimal \textit{RMSE} was applied to experiments. Additionally, in our proposed method there are four hyper-parameters: $\rho$ and $\theta$ (the learning rates in NC-ADMM and ADM), $\epsilon$ (the small positive threshold mentioned in section IV), $\tau$ (finding range in algorithm \ref{alg:Projection1}). We set $\rho=0.001, \theta=0.01$ and $\tau=10^7$ for all experiments. Hyperparameters were selected by ten-fold cross validation (CV) procedure. The value of the hyperparameters with highest accuracy were identified. Experiments with both the proposed algorithm and SDBC were repeated for 20 times with the same experimental and parameter settings. Results shown with these two methods in this paper are mean of experimental outputs followed by the standard deviation. 

\section{Experimental Results}
Consolidated and detailed performance of the different techniques over all synthetic and real data are listed in Table \ref{tab:syntable} and Table \ref{tab:realtable}, respectively. Bar-plots in Figure \ref{fig:Syn and Real result} show respectively \textit{AUC} value (\textit{AUC}) for the outcome and protected attribute, mean difference (\textit{MD}), mean balance residual difference (\textit{BR}) and impact rank ratio (\textit{IRR}) for all techniques over both synthetic (Fig.\ref{fig:Syn and Real result} (a)-(d)) and real data (Fig.\ref{fig:Syn and Real result} (e)-(h)). Experiments using SSEM and SSBR were repeated with various strata, i.e. 4 parallel trials with strata numbers from 2 to 5, respectively. The best experimental result was selected over four strata and shown together with our algorithm in Fig.\ref{fig:Syn and Real result}.

\begin{figure}[h]
\centering
        \includegraphics[width=0.35\textwidth]{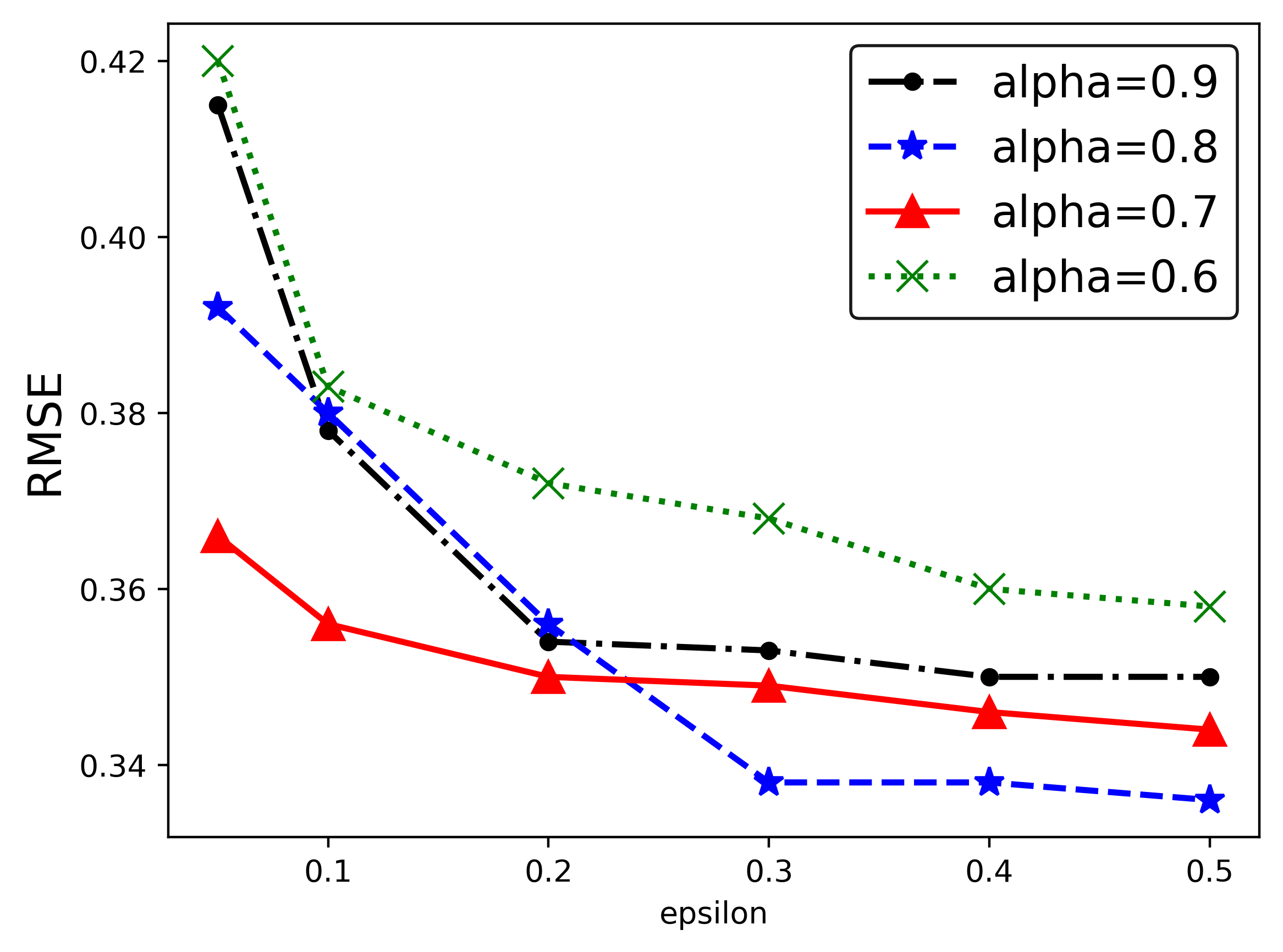}
        \caption{$\epsilon$-\textit{RMSE}, Synthetic Data}
    \label{fig:synRMSE-e}
\end{figure}

\begin{figure}[h]
\centering
    \begin{subfigure}[b]{0.24\textwidth}
        \includegraphics[width=\textwidth]{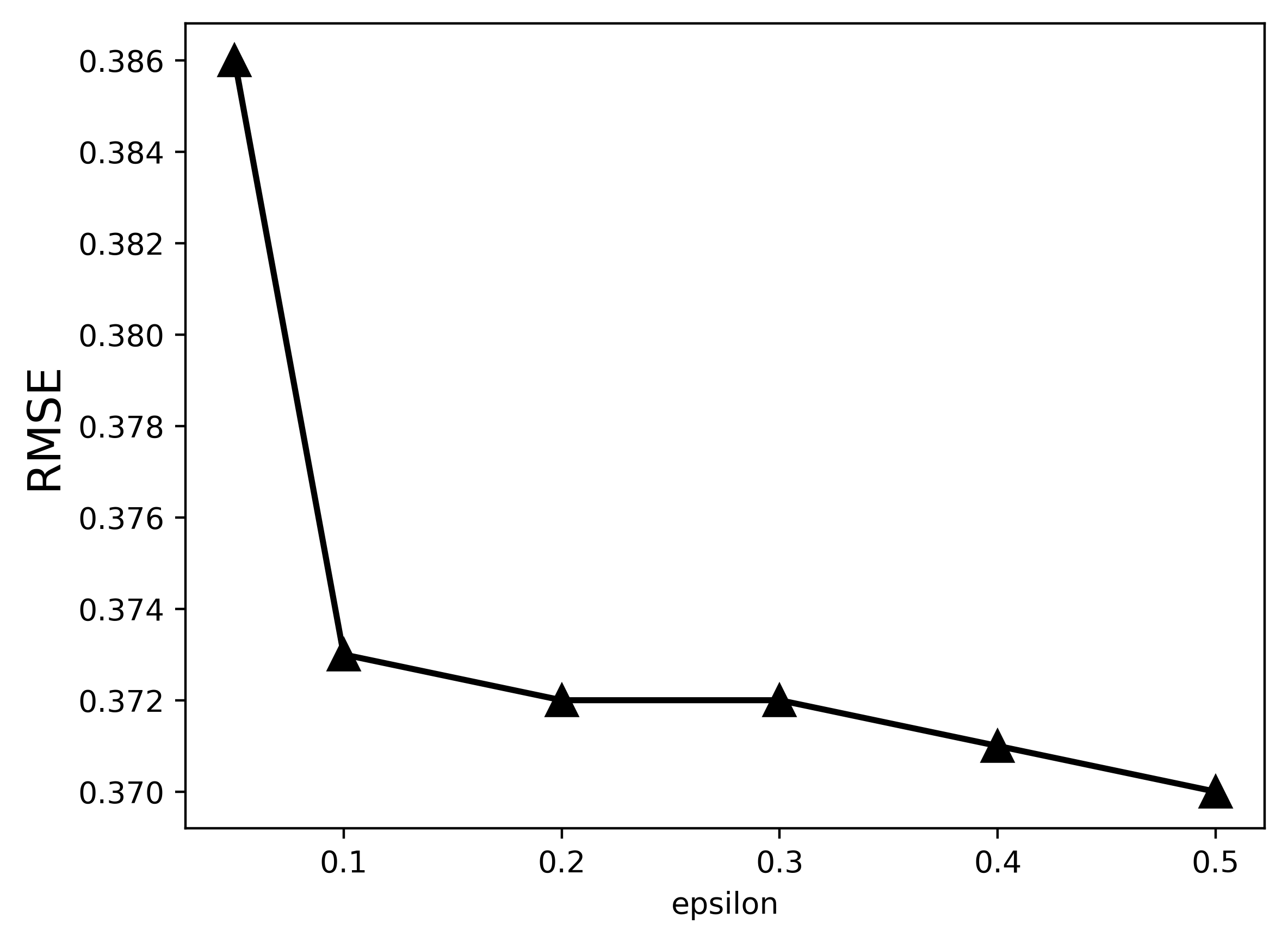}
        \caption{$\epsilon$-\textit{RMSE}, Crime}
    \end{subfigure}
    \begin{subfigure}[b]{0.24\textwidth}
        \includegraphics[width=\textwidth]{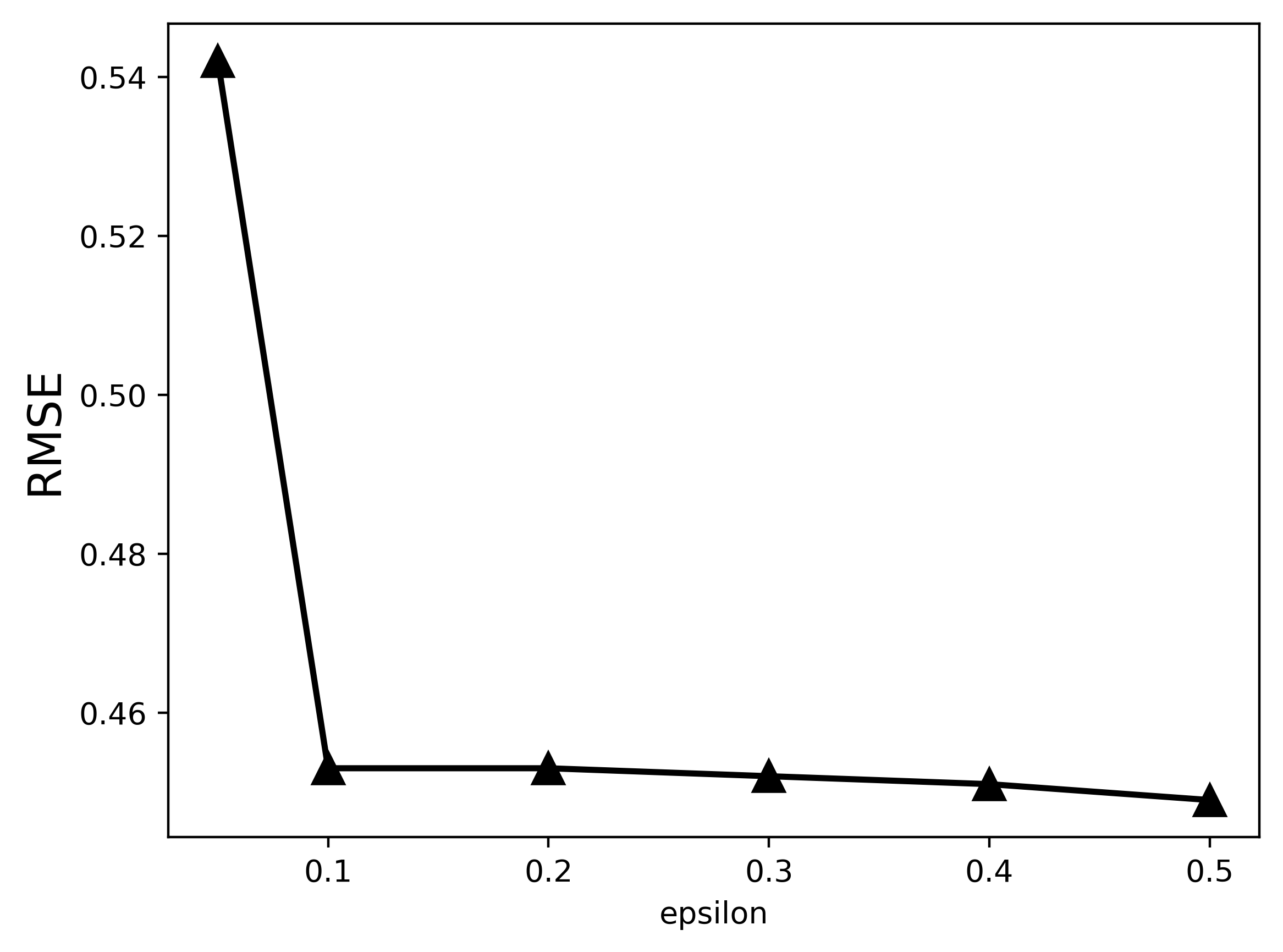}
        \caption{$\epsilon$-\textit{RMSE}, Income}
    \end{subfigure}
    \begin{subfigure}[b]{0.24\textwidth}
        \includegraphics[width=\textwidth]{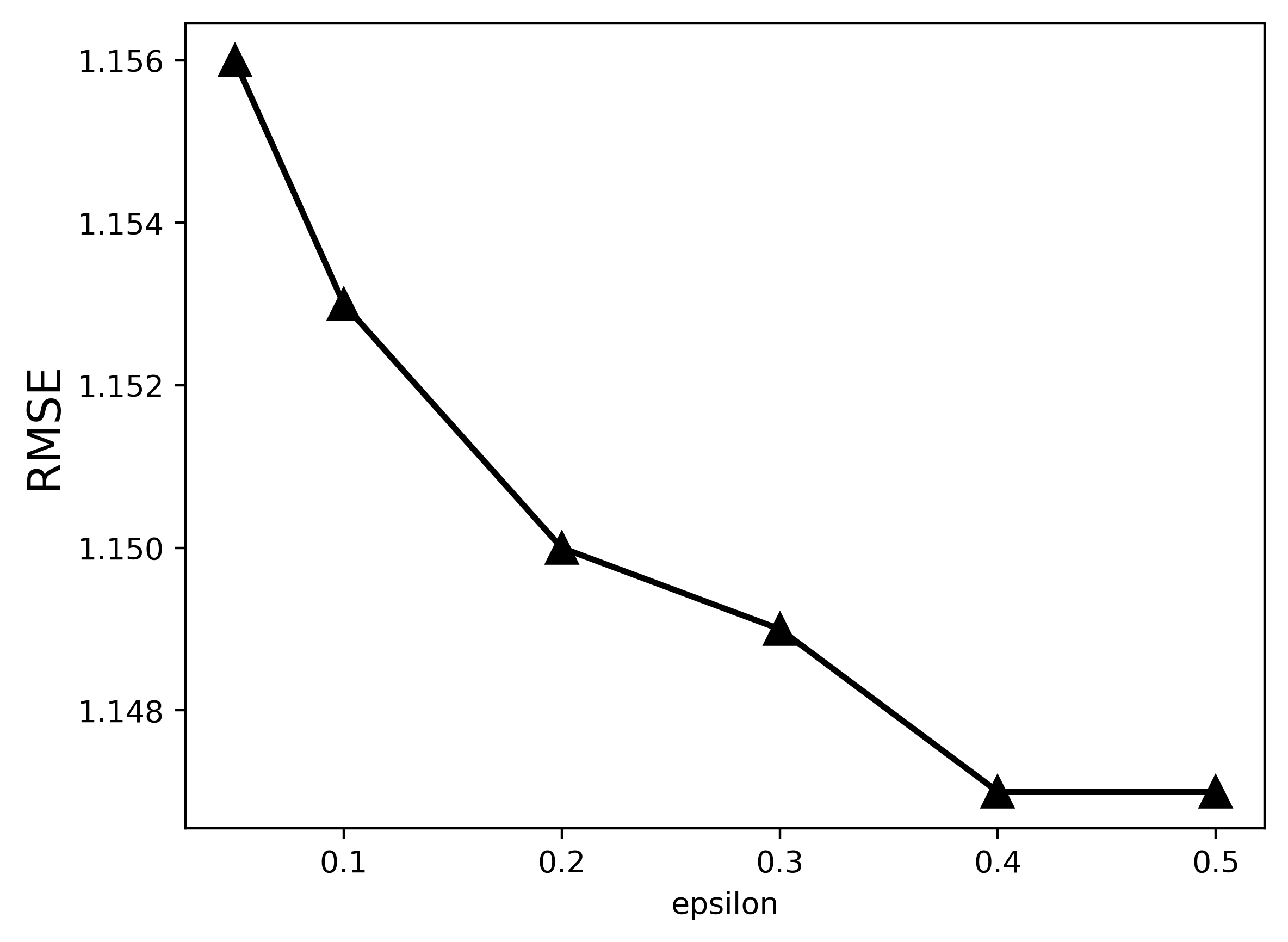}
        \caption{$\epsilon$-\textit{RMSE}, Wine}
    \end{subfigure}
    \begin{subfigure}[b]{0.24\textwidth}
        \includegraphics[width=\textwidth]{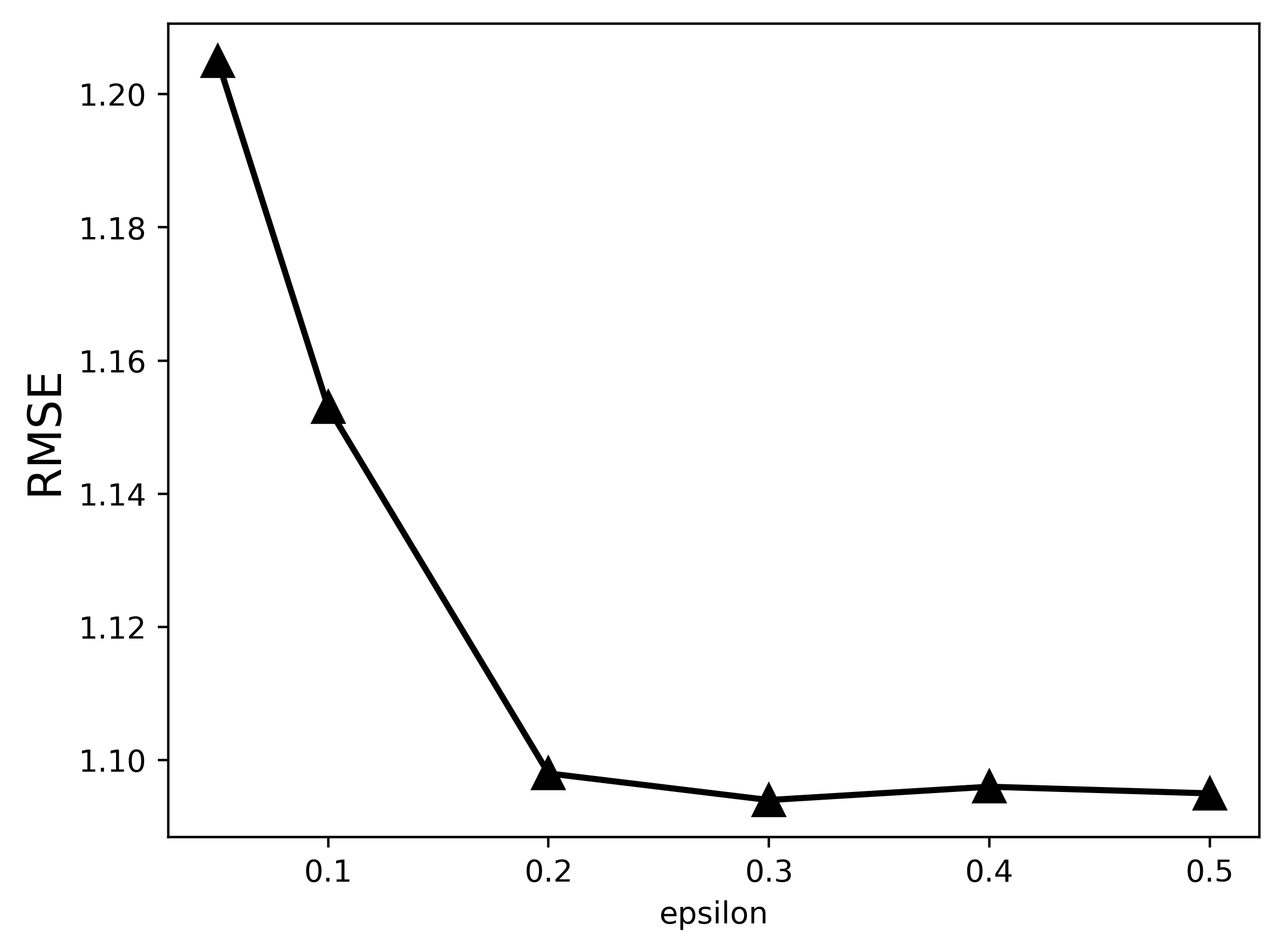}
        \caption{$\epsilon$-\textit{RMSE}, Student}
    \end{subfigure}
    \caption{$\epsilon$-\textit{RMSE} for real data.}
    \label{fig:realRMSE-e}
\end{figure}

Experiment result for synthetic data demonstrated our algorithm (noted as ``Rank" in figures) out-performed than other baseline methods in terms of controlling biases. It restricted \textit{AUC} and \textit{MD} in a soft fair level (i.e. near 0.5 and 0, respectively, Fig.\ref{fig:Syn and Real result} (a) and (b)) and hence increased \textit{IRR} above the boundary of bias level of 0.8 (Fig.\ref{fig:Syn and Real result} (d)). Comparison with baselines, in general, Fig.\ref{fig:Syn and Real result} (c) demonstrated our methods gave the smallest absolute \textit{BR} as well. The SSBR technique provided greater control on \textit{MD} and \textit{BR} over synthetic data (see Fig.\ref{fig:Syn and Real result} (b) and (c)). Unfortunately, SSBR gave contrary results (Fig.\ref{fig:Syn and Real result} (a) and (d)) on restricting \textit{AUC} and increasing \textit{IRR} in comparison using synthetic data. Besides, our result with real data illustrated the proposed algorithm, in contrast to other baseline methods, efficiently controlled the dependency of protected attribute on predictions.
Figure \ref{fig:Syn and Real result} (e) to (h) are experimental results for all evaluation metrics applied with four real datasets. Again, our proposed algorithm performed the best by restricting \textit{AUC} regarding with the protected attribute and predictive outcome to a fair level, minimizing the mean outcome difference and mean balanced residual difference and enhancing the impact rank ratio beyond the the boundary of 0.8 and thus gives bias-free predictions.

\begin{figure}[h]
\vspace{-5mm}
\centering
        \includegraphics[width=0.27\textwidth]{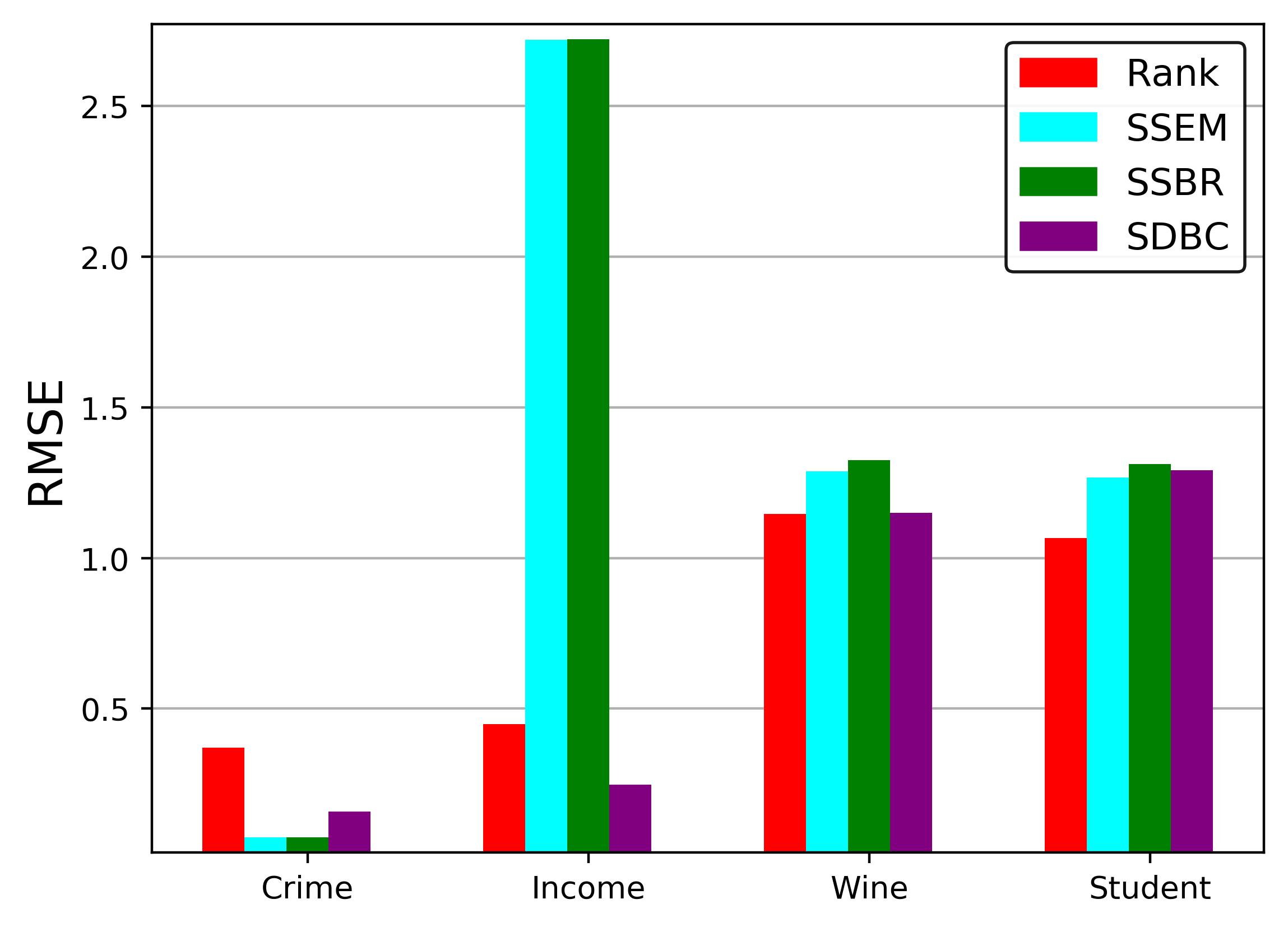}
    \caption{\textit{RMSE} comparison for real data.}
    \label{fig:RMSE}
\end{figure}

Since there is a trade-off between fairness and accuracy, we took \textit{RMSE} into consideration as well (see Fig.\ref{fig:synRMSE-e} to Fig. \ref{fig:RMSE}). Fig.\ref{fig:synRMSE-e} and Fig. \ref{fig:realRMSE-e} show the relation between \textit{RMSE} as a function of the user-defined fairness level $\epsilon$ (smaller is more fair) for both synthetic and real data, respectively. Lower \textit{RMSE} indicates higher prediction accuracy, which may sacrifices prediction fairness as a penalty. Compared with baseline methods (Fig.\ref{fig:RMSE}), our proposed method displayed relative high prediction accuracy with real data.

It is reasonable that SSEM occasionally returns better results than the proposed algorithm. Since SSEM is highly dependent on strata number and a stratum may only have one single partition, this may lead to unstable performance and higher \textit{RMSE}. Additionally, it is interesting that SDBC works well in controlling \textit{AUC} and \textit{MD}. Nevertheless, it gives unexpected performance on \textit{BR}. This may result from the protected attribute is binary, but not continuous. One should note strict covariance between the protected attribute and prediction is not sufficient to obtain independence.

\section{Conclusion}
In this paper, we introduced a novel fairness learning multi-task regression model with rank based non-convex constraint and it efficiently controlled data bias by removing the dependency effect of the protected attribute on predictions across different tasks. In the framework of NC-ADMM, two projected algorithms were proposed to restrict Mann Whitney \textit{U} and further sum rank of the protected attribute to a reasonable boundary region. Experimental results on both synthetic data and real data indicate the proposed algorithm out-performed than traditional baseline methods.

\section*{Acknowledgments}
This work has been supported by the US National Science Foundation under grants IIS-1815696 and IIS-1750911.

\end{document}